\pdfoutput=1
\documentclass{article}

\usepackage[numbers,sort&compress]{natbib}
\usepackage[utf8]{inputenc}
\usepackage[T1]{fontenc}
\usepackage{url}
\usepackage{microtype}
\usepackage{graphicx}
\usepackage{subcaption}
\usepackage{booktabs}
\usepackage{pdfpages}
\usepackage{nicefrac} 
\usepackage{hyperref}
\usepackage{stfloats}
\usepackage{xcolor}
\usepackage{algorithm}
\usepackage{algorithmic}
\usepackage{varwidth}
\usepackage{wrapfig}

\usepackage{amsmath}
\usepackage{amssymb}
\usepackage{mathtools}
\usepackage{amsthm}
\usepackage{amsfonts}
\usepackage{bm}

\usepackage[capitalize,noabbrev,nameinlink]{cleveref}
\DeclareRobustCommand{\abbrevcrefs}{%
\crefname{algorithm}{Alg.}{Algs.}%
}
\DeclareRobustCommand{\cshref}[1]{{\abbrevcrefs\cref{#1}}}

\usepackage{csquotes}
\usepackage{placeins}
\PassOptionsToPackage{hyphens}{url}

\usepackage{adjustbox}
\usepackage{pgf}
\usepackage{pgfplots}
\usepgfplotslibrary{fillbetween}
\usepgfplotslibrary{external}
\usetikzlibrary{backgrounds,calc,arrows.meta,positioning, automata,external}
\usepackage{tikzscale}

\usepackage{color, colortbl}
\usepackage[shortcuts]{extdash}

\usepackage{enumitem}
\usepackage{iftex}
\usepackage{rotating}

\usepackage{svg}

\ifluatex
\else
\DeclareUnicodeCharacter{2212}{−}
\fi

\usepgfplotslibrary{groupplots,dateplot}
\usepgfplotslibrary{fillbetween}
\usepgfplotslibrary{external}
\usetikzlibrary{backgrounds,calc,arrows.meta,positioning,automata,external}
\usetikzlibrary{patterns,shapes.arrows}
\pgfplotsset{compat=1.16}

\newcommand{\neurips}[1]{#1}

\theoremstyle{plain}
\newtheorem{theorem}{Theorem}[section]

\newtheorem{lemma}[theorem]{Lemma}

\theoremstyle{definition}
\newtheorem{definition}[theorem]{Definition}

\theoremstyle{remark}

\DeclareUnicodeCharacter{3B1}{\ensuremath{\alpha}}

\DeclarePairedDelimiter{\ceil}{\lceil}{\rceil}

\usepackage[accepted]{mlsys2025}

\begin{document}

\twocolumn[
\mlsystitle{On Distributed Larger-Than-Memory Subset Selection With Pairwise Submodular Functions}

\begin{mlsysauthorlist}
\mlsysauthor{Maximilian Böther}{ethgoog}
\mlsysauthor{Abraham Sebastian}{goog}
\mlsysauthor{Pranjal Awasthi}{goog}
\mlsysauthor{Ana Klimovic}{eth}
\mlsysauthor{Srikumar Ramalingam}{goog}
\end{mlsysauthorlist}

\mlsysaffiliation{ethgoog}{ETH Zurich, Switzerland. Work partially done while author was interning at Google}
\mlsysaffiliation{eth}{ETH Zurich, Switzerland}
\mlsysaffiliation{goog}{Google, USA}

\mlsyscorrespondingauthor{Maximilian Böther}{mboether@inf.ethz.ch}
\mlsyscorrespondingauthor{Srikumar Ramalingam}{rsrikumar@google.com}

\mlsyskeywords{subset selection, submodularity, distributed algorithms}

\vskip 0.3in

\begin{abstract}
Modern datasets span  billions of samples, making training on all available data infeasible.
Selecting a high quality subset helps in reducing training costs and enhancing model quality.
Submodularity, a discrete analogue of convexity, is commonly used for solving such subset selection problems. 
However, existing algorithms for optimizing submodular functions are sequential, and the prior distributed methods require at least one central machine to fit the target subset in DRAM.
At billion datapoint scale, even the subset may not fit a single machine, and the sequential algorithms are prohibitively slow.
In this paper, we relax the requirement of having a central machine for the target subset by proposing a novel distributed bounding algorithm with provable approximation guarantees.
The algorithm iteratively bounds the minimum and maximum utility values to select high quality points and discard the unimportant ones. When bounding does not find the complete subset, we use a multi-round, partition-based distributed greedy algorithm to identify the remaining subset.
We discuss how to implement these algorithms in a distributed data processing framework and empirically analyze different configurations.
We find high quality subsets on CIFAR-100 and ImageNet with marginal or no loss in quality compared to centralized methods, and scale to a dataset with 13 billion points.

\end{abstract}
]

\printAffiliationsAndNotice{}
\section{Introduction}\label{sec:intro}
The increasing volume of collected data requires identifying highly informative subsets of features or datasets to cost-effectively train high-quality models~\cite{Bhardwaj2022Ekya}.
For instance, vision datasets have scaled from ImageNet's 1.2\,M samples~\cite{Deng2009ImageNet} to LAION's 5\,B samples~\cite{Schuhmann2022LAION}.
A single autonomous vehicle collects terabytes of sensor data daily~\cite{Kazhamiaka2021AV} and recent language/vision models pretrain on billions of examples from books and big webpage collections~\cite{Gao2020Pile, Shen2023SlimPajama}.
The ever-increasing scale of datasets makes it challenging and costly to train ML models on all available data. Selecting high quality samples also improves model quality compared to training with larger, less informative datasets. Subset selection is also crucial in applications beyond ML model training, such as feature selection, dictionary learning, and numerous compressed sensing applications~\cite{Krause2010,Das2011}.

Subset selection is a well studied problem with many competing algorithms that rely on submodularity, coresets, and other clustering-based methods.
Various greedy algorithms offer strong approximation guarantees but are inherently sequential and centralized~\cite{Nemhauser1978Submod}.
Existing distributed methods typically partition the dataset, run the greedy algorithm on each partition, and use the greedy algorithm again on the union of the subsets from the  partitions~\cite{Mirzasoleiman2016Greedi, Barbosa2015Greedy}.
This final step, mainly used to achieve approximation guarantees, needs a machine that holds the results from all partitions.
This can require unfeasible terabytes of DRAM.
We are not aware of  prior methods that are tested on  massive datasets. 

We present a novel bounding algorithm that iteratively tightens the maximum and minimum utilities of the individual points.
This allows to identify high-utility points and discard less informative ones in a distributed fashion. 
This algorithm is highly parallelizable and can be implemented in distributed data processing frameworks, such as Apache Beam~\cite{Akidau2015Beam}.
If the bounding algorithm does not compute the complete subset, we employ a distributed, multi-round, partition-based greedy algorithm to achieve high-quality subsets.
With bounding and multi-round partition-based optimization, we are able to select high quality subsets that achieve similar quality to those obtained by centralized algorithms.
\emph{Neither our bounding method nor the distributed greedy algorithm requires a central machine with enough DRAM to hold the final subset.}

Among many competing subset selection methods, we opt for pairwise submodular functions for several reasons.
First, submodular functions have been shown to produce state-of-the-art results in a variety of applications (c.f.~\Cref{sec:prior}). Second, they allow us to use a graph for encoding point similarities without needing to consider more complex interactions, i.e., hyper-edges or higher-order terms which cannot be constructed efficiently~\cite{Ramalingam20171}. 
Many existing techniques such as the generalized graph cut function in~\cite{Iyer2020, Kothawade2021}, k-medioids in~\cite{Park2009}, and prototype selection as expressed in Eqn. (6) in~\cite{Kim2016}, and the chosen submodular utility in GIST~\cite{fahrbach2025gist} can all be seen as the maximization of pairwise submodular functions.
We make the following contributions.
\begin{itemize}[leftmargin=*, nosep, topsep=0pt]
\item We design a highly-parallelizable bounding algorithm that, by continuously adjusting maximum and minimum utility values, identifies high utility points to expand the subset and discards uninformative ones to reduce the ground set.
We also show approximation guarantees. 
\item The bounding algorithm does not always find a complete subset.
To compute the remaining subset, when necessary, we introduce a distributed multi-round partition-based algorithm that empirically achieves similar quality results as the centralized greedy algorithm. 
\item We show that our distributed methods lead to similar quality results as centralized methods on the CIFAR-100 (50\,k samples) and ImageNet (1.2\,M samples) datasets~\cite{Krizhevsky2009CIFAR, Deng2009ImageNet}. 
We also study scalability on a 13\,B Perturbed-ImageNet dataset that we generate from ImageNet.
\end{itemize}

\section{Prior Work}\label{sec:prior}

\textbf{Submodular subset selection.}  Several discrete problems, such as cut functions, coverage, and entropy maximization, can be formulated as the minimization or maximization of submodular functions.
Many subset selection tasks can be modeled as submodular maximization problems and applications of submodularity include data selection, feature engineering, sensor placement, and influence maximization~\cite{Carbonell98,Simon2007,krause2014submodular,Wei2015,lin-bilmes-2011-class,kaushal2019learning,Badanidiyuru2014,Kim2016,Prasad2014, golovin2010adaptive,manjunatha2018class,Ramalingam2021Submod,Kothawade2021,Kaushal2021}.
While submodular functions can be minimized in polynomial time, their maximization is NP-hard.
Efficient greedy maximization algorithms with approximation guarantees exist~\cite{Nemhauser1978Submod}.
Even after four decades, this simple greedy method is the gold standard for centralized submodular maximization \neurips{under cardinality constraints.}

\textbf{Distributed algorithms.} 
\citet{Mirzasoleiman2016Greedi} present the distributed \textsc{GreeDi} algorithm for maximizing monotone submodular functions under a cardinality constraint.
The data is partitioned arbitrarily across machines, and each machine runs the centralized greedy algorithm on its assigned data.
Then, the greedy algorithm is run again on the union of the results. 
This can be implemented in the distributed MapReduce framework~\cite{Dean2008}.
\citet{Barbosa2015Greedy} extend \textsc{GreeDi} to \textsc{RandGreeDi} by changing the way the data is assigned to each machine, leading to constant-factor worst case approximation guarantees.

\citet{Kumar2015} develop \textsc{Sample\&Prune}, a MapReduce algorithm for maximizing a monotone submodular function under a cardinality or matroid constraint. 
\textsc{Sample\&Prune} has a constant approximation bound but assumes that the memory per machine is \(\mathcal{O}\left(k n^{\delta}\right)\), where $k$ is the cardinality of the final subset, \(n\) is the number of input points, and $\delta > 0$.
\citet{Liu2018} discuss a MapReduce algorithm with constant number of rounds while assuming that there exists a central machine with \(\mathcal{O}(\sqrt{nk} \log\left(k\right))\) memory. \citet{Barbosa2016Framework} discuss another MapReduce algorithm with a constant number of rounds and strong guarantees, while assuming that a single machine should fit the final solution with \(\mathcal{O}(n^{1 - 2c})\) memory, where $0<c<0.5$.

Subset selection is related to coresets, which the idea is that the solution to an optimization on this coreset closely matches the solution from the entire dataset~\cite{Piotr2014,MirrokniZ15}.
To reduce the memory footprint on a single machine, sketching algorithms have been explored for distributed submodular optimization~\cite{Bateni2017,Bateni2018}.
In the context of clustering, parallel algorithms have also been studied for the \(k\)-center objective~\cite{ene2011fast, im2015fast, malkomes2015fast, mcclintock2016efficient,ramalingam2023weighted}. 
Iterative grow-and-shrink approaches for submodular optimization have also been studied in other contexts~\cite{Zhou2017Scaling, Buchbinder2012Submod}.

\begin{table}
\centering
\caption{{\it Maximum sizes of datasets considered in prior works on (distributed) submodular subset selection.}}
\label{tab:scale-comparison}
\small
\adjustbox{max width=\linewidth}{%
\begin{tabular}{l|r}
\multicolumn{1}{c|}{work}       & \multicolumn{1}{c}{max. subset / ground set size}  \\ 
\hline
\citet{Barbosa2015Greedy}       & 120 /~1\,M                                         \\
\citet{Mirzasoleiman2016Greedi} & 64 /~80\,M                                         \\
\citet{Ramalingam2021Submod}    & 700\,k /~1.2\,M                                   \\
\citet{Kumar2015}               & 500 /~1\,M                                         \\
this paper                      & 6.5\,B /~13\,B                                    
\end{tabular}
}
\vspace{-0.5cm}
\end{table}

\textbf{Limitations of prior methods.}
We see two main limitations: (1) almost all methods rely on having a central machine that holds the entire subset to show strong theoretical guarantees, and (2) validation on really large datasets has received limited to no attention, as shown in~\Cref{tab:scale-comparison}.
We take a practical \emph{systems approach} and demonstrate algorithms scaling to massive datasets with billions of points, by leveraging a subclass of submodular functions, which is effective on a large class of subset selection problems. 

\section{Centralized Subset Selection}\label{sec:centralized-algo}

We start by defining the two fundamental concepts of submodularity and monotonicity.
\begin{definition}[Submodularity] 
Let \(\Omega\) be a finite set.
A set function \(F:2^{\Omega} \rightarrow \mathbb{R}\) is \emph{submodular} if for all \(A,B \subseteq \Omega\) with \(B \subseteq A\) and \(e \in \Omega \setminus A\),
\(F(A \cup \{e\}) - F(A) \le F(B \cup \{e\}) - F(B)\).
\label{defn:submodularity}
\end{definition}
This property is also referred to as \emph{diminishing returns} since the gain diminishes as we add elements.
\begin{definition}[Monotonicity] 
A set function \(F\) is \emph{monotonically non-decreasing} if for  $B \subseteq A$, $F(B) \le F(A)$.
\label{defn:monotonicity}
\end{definition}
Given a ground set \(V\), the goal of centralized submodular subset selection is to find a subset \(S \subseteq V\) of size \(k \leq |V|\) that maximizes a submodular and monotone non-decreasing set function  \(f: 2^V \rightarrow \mathbb{R}\).
We want to determine \(S = \arg\max_{S' \text{ with } |S'| = k} f(S') \).
Since \(f\) is submodular, we can use an efficient greedy algorithm to compute a subset \(S^\prime\) with constant approximation guarantee \(f\left( S^\prime \right) \geq \left( 1 - \frac{1}{e} \right) f\left( S_{\text{OPT}} \right)\)~\cite{Nemhauser1978Submod}.
As outlined in~\Cref{alg:greedy-hl}, the greedy algorithm repeatedly chooses the data point \(v\in V\setminus S\) that maximizes the marginal gain upon adding the element to the subset.

\begin{algorithm}[H]
  \caption{Centralized greedy algorithm\\ for computing subset of size \(k\in\mathbb{N}\).}
  \label{alg:greedy-hl}
\begin{algorithmic}[1]
  \STATE \(S \gets \emptyset\)
  \WHILE{\(|S| < k\)}
  \STATE \( S \gets S \cup \left\{ \arg\max_{s\in V \setminus S} f\left( S \cup \left\{s\right\} \right) - f\left(S\right)\right\} \) %
  \ENDWHILE
\end{algorithmic}
\end{algorithm}

Let \(\alpha, \beta\) be balancing parameters.
Let \(E\) denote the edges of a nearest neighbor graph, i.e., a graph where connected vertices (data points) \(v_1\) and \(v_2\) are neighbors with edge weight (similarity) \(s(v_1, v_2)\).
Additionally, let \(u(v)\) denote the utility of \(v\).
We focus on the class of pairwise submodular functions \(f(S) = \alpha \sum_{v \in S} u(v) - \beta \sum_{(v_1, v_2) \in E;v_1, v_2 \in S} s\left(v_1, v_2\right)\), used in subset selection~\cite{Kim2016,Ramalingam2021Submod}, graph cuts~\cite{Iyer2020,Kothawade2021,Kolmogorov2004}, and diversification~\cite{Borodin2017,Bhaskara2016Linear, Gollapudi2009}.
This class \textit{(i)} allows us to balance and utility and diversity and \textit{(ii)} is more tractable than higher order functions. 
For billion-scale datasets, it is not feasible to efficiently compute interactions of third order or higher.
Second order interactions can be computed with careful engineering, approximations, and optimizations.

We start by showing that such functions are always submodular and discuss conditions for monotonicity.
Let \(f\) be  a function of this class and assume \((v_1, v_2) \in E;v_1, v_2 \in S\), \(s(v_1, v_2) \geq 0\).
Further, consider two sets $A$ and $B$ such that $B \subseteq A$ and let $e \subseteq V\setminus A$.
Then, $(f(A \cup e) - f(A)) - (f(B \cup e) - f(B)) = \beta \sum_{j \in B} s(e, j) - \beta \sum_{j \in A} s(e, j)$.
Since $\beta \geq 0$, $\forall a,b$, $s(a, b) \geq 0$, and $B \subseteq A$, we can see the diminishing returns property as $(f(A \cup e) - f(A)) - (f(B \cup e) - f(B)) \leq 0$.
The function is always submodular. 

The monotonicity condition holds true when for all \(v \in S\), we have $\alpha u(v) \geq \beta \sum_{(v_1, v_2) \in E;v_1, v_2 \in S} s\left(v_1, v_2\right)$ for all points $v$.
While this condition holds true on most datasets, in cases where this condition does not hold true, we can easily add a constant offset $\delta$ to the unary terms of all data points and thereby ensure monotonicity (c.f.~\Cref{lb.monotonic}). In this case, the standard approximation guarantee of $f(S)\ge (1-\frac{1}{e})f(S_{\text{OPT}})$ becomes $f(S)+k\delta \ge (1-\frac{1}{e})(f(S_{\text{OPT}} + k\delta)$.

\begin{algorithm}[H]
   \caption{Centralized implementation of~\Cref{alg:greedy-hl} using a priority queue.}
   \label{alg:greedy-central}
\begin{algorithmic}[1]
   \STATE \(q \gets \text{new priority queue} \), \(S \gets \emptyset\)
   \STATE \(\forall v \in V: \text{Insert } v \text{ into } q \text{ with weight } u(v) \)
   \WHILE{\(|S| < k\)}
   \STATE \(v_1 \gets q.\mathsf{popmax}()\)
   \FOR{\textbf{each} \( v_2 \in p \text{ with } s(v_1, v_2) > 0 \)}
   \STATE \( q.\mathsf{decrease\_weight\_by}\left(v_2, \beta/\alpha \cdot s\left(v_1, v_2\right)\right) \)
   \ENDFOR
   \STATE \( S \gets S \cup \left\{ v_1 \right\} \)
   \ENDWHILE
\end{algorithmic}
\end{algorithm}

These functions enables implementing the algorithm using a priority queue as shown in~\Cref{alg:greedy-central}. 
We initialize all of \(V\) into the queue with their utility scores as priority. Then, we repeatedly pop the element \(v_1\) with highest priority, add it to the subset, and update the priority of all neighboring elements \(v_2\) where \((v_1, v_2) \in E\).
Since we only update the priorities of the neighbors of \(v_1\), this enables to efficiently build up \(S\) without the need to repeatedly calculate \(f\left(S \cup \{v\}\right)\) for all \(v\).
We continue to iterate until we build a subset of cardinality $k$.

\textbf{Scaling challenges.}
We see two major roadblocks. First, the algorithm is inherently sequential and relies on identifying the element that maximizes the marginal gain over the entire dataset at each step.
The computation of the marginal gain with respect to the entire dataset makes it difficult to parallelize. Second, the implementation requires large amounts of DRAM to load the keys and similarity values. For datasets with billions of points, this requires more DRAM than commodity VMs or even high-end servers offer. For example, storing 5\,billion 64-bit keys and values in the priority queue, and keeping track of 10 nearest neighbors with 64-bit IDs and distances requires 880\,GB of memory.

\textbf{Related optimizations.}
Optimized centralized optimization algorithms for general submodular functions, such as lazy greedy~\cite{Minoux1978AcceleratedGA}, stochastic greedy~\cite{MirzasoleimanBKVK14}, and threshold greedy~\cite{Badanidiyuru2014SODA}, are orthogonal to the challenging of scaling to datasets with billions of points, and might even be disadvantageous for two reasons.
First, we consider the special case of pairwise submodular functions, where weights can be updated efficiently by looking at a small number of neighbors (10 in~\Cref{sec:evaluation}).
Second, using lazy greedy without updating priorities on neighbor selection would make marginal gain computations more expensive later, as they would rely on all points in the current subset.
Nevertheless, in the following, we propose techniques to scale out the selection using \emph{any centralized version} of the algorithm.

\section{Distributed Submodular Subset Selection}\label{sec:algorithm}

Our algorithm has two components, which are executed sequentially: (1) bounding, and (2) distributed greedy optimization. 
First, the bounding algorithm alternates between (1) removing points from the ground set that are less likely to be in the subset and (2) adding points in the subset that are more likely to be in the optimum set, using their minimum and maximum utility. We can implement the bounding algorithm using  data processing frameworks with a large number of machines without the need to store the final subset in a single machine.  We propose both exact (\Cref{subsec:bounding}) and approximate bounding (\Cref{subsec:approx-bounding}) algorithms, with approximation guarantees (\Cref{subsec:theory}).
Second, as the bounding algorithm is not guaranteed to find the complete subset in all cases, we use a distributed greedy algorithm  to find the remaining points if necessary (\cref{subsec:distri-algo}). 

\subsection{Exact Bounding}\label{subsec:bounding}
\begin{figure*}
    \centering
    \adjustbox{trim=1.8cm 3.2cm 3cm 0cm}{%
        \includesvg[width=\textwidth]{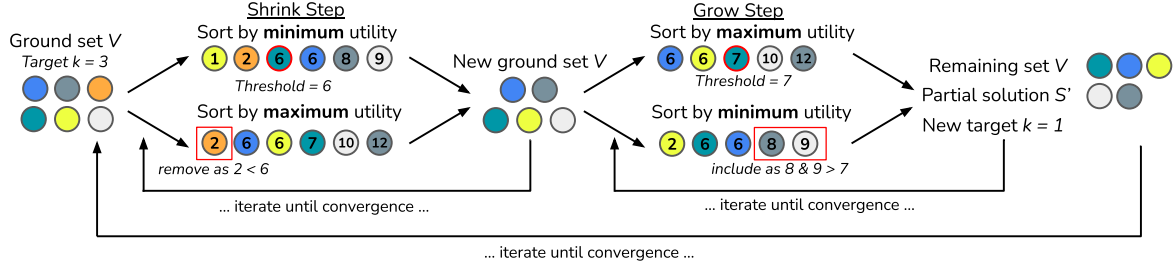}
    }
    \caption{{\it Visualization of distributed bounding when finding a 50\,\% subset for 6 data points.}} 
    \label{fig:bounding-viz}
\end{figure*}

In the exact bounding algorithm, the addition of points to the subset and removal of points from the ground set are guaranteed to preserve the subset quality. 
The bounding algorithm uses two metrics --- the minimum and maximum utilities of a point --- that depends on the current partial subset \(S'\) and the remaining set \(V\).
\begin{definition}[Minimum Utility]\label{def:min_util}
For a point \(v_1\in V\), its \emph{minimum utility} is its utility considering all its neighbors that have not been discarded: \(U_{\text{min}}(v_1) = u(v_1) - \frac{\beta}{\alpha} \sum_{v_2 \in V \cup S' \wedge (v_1, v_2) \in E} s(v_1, v_2) \).
\end{definition}
\begin{definition}[Maximum Utility]
For a point \(v_1\in V\), its \emph{maximum utility} is its utility when only considering the neighbors in the partial subset \(S'\): \(U_{\text{max}}(v_1) = u(v_1)  - \frac{\beta}{\alpha} \sum_{v_2\in S' \wedge (v_1, v_2) \in E} s(v_1, v_2)  \).
\end{definition}
Before we present the bounding algorithm, we introduce two basic blocks for growing the current subset \(S'\), and shrinking the remaining set \(V\) based on the minimum and maximum utilities of the points.
Let \(U_{\text{min}}^i\) and \(U_{\text{max}}^i\) denote the \(i\)-th largest minimum and maximum utility points, respectively.

\textbf{Grow and Shrink Steps.} Let \(S^*\) denote the optimal solution.
The idea behind growth is to identify points with minimum utility higher than the \(k\)-th largest maximum utility and include these in the subset, where \(k\) is the number of points we still need to find.
{%
\setlength{\parskip}{0pt}
The following lemma and~\Cref{alg:grow-bounding} summarize this. 
\begin{lemma}
    For \(v\in V\), if \(U_{\text{min}}(v) > U_{\text{max}}^k\), then \(v\in S^*\).
    \label{lemma.growth}
\end{lemma}

\begin{algorithm}[H]
   \caption{\(\mathsf{Grow}(V, S, k)\)}
   \label{alg:grow-bounding}
\begin{algorithmic}
    \STATE \(\forall v\in V\): Compute \(U_{\text{min}}(v)\) and \(U_{\text{max}}(v)\).
    \STATE \( \text{Threshold} \gets U_{\text{max}}^k \)
    \STATE \( S' \gets \left\{  v \in V \mid U_{\text{min}}(v) > \text{Threshold} \right\} \)
    \STATE \textbf{return} \(V \setminus S', S' \cup S, k - |S'|\)
\end{algorithmic}
\end{algorithm}

Analogously, as per the following lemma and~\Cref{alg:reduce-bounding}, we eliminate points with very low maximum utility, i.e., points whose maximum utility is lower than the \(k\)-th largest minimum utility. 
\begin{lemma}
    For \(v\in V\), if \(U_{\text{max}}(v) < U_{\text{min}}^k\), then \(v\notin S^*\).
\end{lemma}
}%
    
\begin{algorithm}[H]
   \caption{\(\mathsf{Shrink}(V, S, k)\)}
   \label{alg:reduce-bounding}
\begin{algorithmic}
    \STATE \(\forall v\in V\): Compute \(U_{\text{min}}(v)\) and \(U_{\text{max}}(v)\).
    \STATE \( \text{Threshold} \gets U_{\text{min}}^k\)
    \STATE \textbf{return} \( \left \{ v \in V \mid U_{\text{max}}(v) \ge \text{Threshold} \right \} \)
\end{algorithmic}
\end{algorithm}

We use these grow and shrink blocks (c.f.~\Cref{alg:basic-bounding}), and repeat each block until convergence before we alternate to the other one.
We may be able to grow more points after repeated shrink steps, or shrink more points after repeated grow steps since the grow step will decrease the maximum utility of some points, and the shrink step will increase the minimum utility of some points. 
We give a visual intuition in~\Cref{fig:bounding-viz}.

\begin{algorithm}[H]
   \caption{\(\mathsf{Bound}(V, k)\)}
   \label{alg:basic-bounding}
\begin{algorithmic}
    \STATE \(S' \gets \emptyset\)
    \REPEAT
    \STATE \textbf{repeat} \(V \gets \mathsf{Shrink}(V, S', k)\) \textbf{until \(V\) converges}
    \STATE \textbf{repeat} \(V, S', k \gets \mathsf{Grow}(V, S', k)\) \textbf{until \(S'\) converges}
    \UNTIL{\(S'\) and \(V\) converge}
    \STATE \textbf{return} \(V, S', k\)
\end{algorithmic}
\end{algorithm}

\subsection{Approximate Bounding}\label{subsec:approx-bounding}
While the exact bounding algorithm can provide optimal subsets, even better than the centralized greedy algorithm, it is not guaranteed to find the complete subset.
In practice, we find that the algorithm converges very quickly yielding very incomplete subsets (\Cref{subsec:eval-bounding}). 
This is due to the strict condition to only discard and add points when we are absolutely certain.
In this subsection, we relax this constraint, and present an approximate bounding algorithm using the notion of expected utility as defined below. 
\begin{definition}[Expected Utility]\label{def:exp_util}
For any point \(v_1\in V\), its \emph{expected utility} considers only a subset of neighbors: \(U_{\text{exp}}\left(v_1\right) = u\left(v_1\right) - \frac{\beta}{\alpha} \sum_{v_2 \in U\left(N_{v_1}\right) \cup S'} s\left(v_1, v_2\right) \), where \(U\left(N_{v_1}\right)\) denotes the set obtained by sampling the neighbors of \(v_1\) from the remaining set, either uniformly or weighted based on the similarity values.
\end{definition}
Note that we always consider all neighbors in the current partial solution \(S'\), and, in our implementation (\Cref{sec:evaluation}), only sample if the number of neighbors in \(S'\) is too small.
With the expected utility, the approximate bounding algorithm works the same as the exact bounding where we replace the minimum with the expected utility. 
Replacing the minimum utility with the expected utility intuitively leads to removal of a larger number of points from the ground set and the addition of a larger number of points in the subset.
While approximate bounding is not guaranteed to find subsets of optimal quality as in the case of exact bounding, we show provable guarantees on the quality of the subset with respect to the centralized greedy solution in the next subsection. 
Note that this algorithm might grow \(S'\) larger than needed, in which case we sample a subset of the correct size uniformly at random.

\subsection{Theoretical Analysis of Approximate Bounding}\label{subsec:theory}
In this section we provide theoretical justification for the approximate bounding procedure.
First notice the by performing {\em exact} bounding followed by running the centralized greedy submodular procedure at the end at least get a $\frac 1 2$-approximation guarantee since the exact bounding procedure never discards elements from the optimal set $S^*$. For this theoretical analysis, we sample the neighborhood uniformly at random. This means for each $v \in V$, each vertex in the neighborhood $N_v$ is chosen independently at random with probability $p$ when calculating $U_{\text{min}}(v)$. 

\begin{theorem}
\label{thm:main}
Let the given input instance be such that each non-zero similarity value lies in the range $[a,b]$. Furthermore, let $\gamma > 1$ be such that at the beginning of the algorithm, for each $v \in V$ it holds that $\frac{U_{\text{max}}(v)}{U_{\text{min}}(v)} \leq \gamma$. If random sampling with probability $p$ is used for approximate bounding then with probability at least $1-|V| e^{-p^2 (1-p)^2 k_g \frac{a^2}{(b-a)^2}}$ the algorithm outputs a set $S$ of size $k$ such that \(f(S) \geq \frac{1}{2\big(1+\gamma(1-p^2) \big)} f(S^*)\), and $k_g$ in the exponent refers to the minimum degree of the graph.
\end{theorem}

The approximation guarantee becomes better as $p$ increases, which is expected. For $p=1$ we recover the $\frac{1}{2}$-guarantee of exact bounding. The guarantee also depends on the ratio of the maximum and the minimum utilities which in turn is related to the balancing parameters \(\alpha, \beta\). In the worst case, when this ratio becomes infinity, we get a vacuous bound which is expected since we use approximate bounding. We refer to~\Cref{thm.proof} for the proof. 

\subsection{Distributed Greedy Algorithm}\label{subsec:distri-algo}
\begin{figure*}
    \centering
    \adjustbox{trim=0cm 4.3cm 4.2cm 0cm}{%
        \includesvg[width=\textwidth]{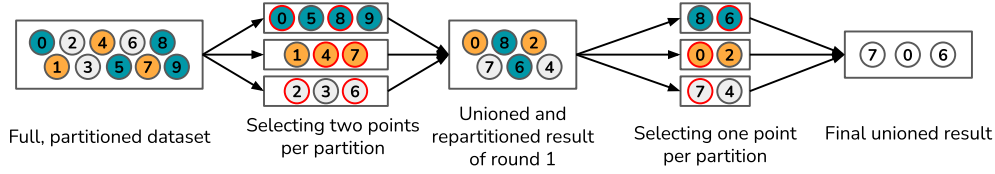}
    }
    \caption{{\it Visualization of the distributed submodular algorithm finding a subset of size 3 out of 10 points using 2 rounds with 3 partitions. The partitioning is given by color,  the selected points per partition are marked with a red border, and the numbers represent IDs.}}
    \vspace{-0.5cm}
    \label{fig:distsub-viz}
\end{figure*}

The bounding algorithm is not always guaranteed to provide the complete subset, and in such cases, we use a partition-based distributed algorithm to compute the rest of the points in the subset. 
Prior distributed algorithms first partition the entire dataset, run centralized greedy in each of these partitions, and typically use a final step where a centralized greedy is run on the union of the individual subsets computed in each of the partitions (c.f.~\Cref{sec:prior}).
However, this final step of running the centralized greedy on the union is infeasible when the size of the subset is large. 
We skip this final subset selection on the union, and instead produce smaller subsets in each of the partitions whose union directly lead to the desired subset.
A union can be implemented without materializing all data in memory by data processing frameworks such as Spark~\cite{Zaharia2016Spark}, Flume~\cite{Chambers2010Flume}, or Beam~\cite{Akidau2015Beam}.
To produce a high quality subset, we employ several rounds where we iterate over finding subsets in different partitions, and partitioning the union of the computed subsets for the next round.

\begin{algorithm}[H]
   \caption{\textit{Distributed greedy using adaptive partitioning over \(m\) machines for \(r\) rounds to find subset \(S\) of size \(k\).}}
   \label{alg:distri}
\begin{algorithmic}
    \STATE \(\text{partition\_cap} \gets \ceil{\frac{|V|}{m}}\)
    \STATE \(V_0 \gets V \)
    \FOR{\(\text{round} = 1 \ldots r\)}
    \STATE \(n_{\text{round}} \gets \Delta(|V|, r, \text{round}, k) \)
    \STATE \(m_{\text{round}} \gets \ceil{\frac{n_{\text{round}}}{\text{partition\_cap}}}\)
    \STATE \(P_{1}, \ldots, P_{m_{\text{round}}} \gets \text{random partition of }  V_{\text{round}-1} \)
    \FOR{\textbf{each partition } \(i = 1\ldots m_{\text{round}}\) \textbf{ in parallel}}
    \STATE \( S_i \gets  \text{Centralized Greedy (\cshref{alg:greedy-central}) on } P_i\)
    \ENDFOR
    \STATE \(V_{\text{round}} \gets \bigcup_{i = 1\ldots m_r} S_i \)
    \ENDFOR
    \STATE \(S \gets \mathsf{subsample}(V_r, k)\) \textbf{if} \(|V_r| > k\) \textbf{else} \(V_r\)
\end{algorithmic}
\end{algorithm}

We show an example in~\Cref{fig:distsub-viz} for the distributed greedy algorithm oulined in~\Cref{alg:distri}.
Given the initial dataset size \(|V|\), the overall number of rounds \(r\), the current round \(\text{round}\), and the target size of the last round \(k\), the function \( \Delta\) gives us the number of data points to keep in this round. %
This can, for example, be a linear interpolation between \(|V|\) at the beginning and \(k\) in the last round.
However, many choices of \(\Delta\) are possible and the only constraint is that \(\Delta(|V|, r, r, k) = k\) to ensure we output a subset of the correct size.

After determining the target size \(n_r\) for the current \(\text{round}\), we determine how many partitions, and therefore nodes, to use this round.
We differentiate two variants of the algorithm.
In \cref{alg:distri}, we show our adaptive partitioning algorithm where we scale the number of partitions based on how many nodes we actually need to fit the current dataset \(V_{\text{round}}\).
The intuition behind adaptive partitioning is that using more partitions will perform worse due to less global information, and hence we want to utilize the minimum number of partitions needed.
We can also disable adaptive partitioning, in which the number of partitions each round \(m_r\) is always equal to the number of machines \(m\) available at the beginning. 
In this paper, we limit ourselves to partitioning uniform at random. 
Then, in parallel on each node, we start the centralized algorithm on its assigned partition, with the target to find a dataset of size \(\ceil{n_r / m_r}\).
We discard any neighborhood relation across partitions while computing the submodular objective. 
After all partitions have been processed, we union the results of each partition. 
We repeat this procedure for \(r\) rounds.
At the end, the resulting dataset \(V_r\) might contain up to \(m_r\) additional points due to rounding. 
Hence, we obtain \(S\) by subsampling \(k\) points from \(V_r\).
In~\cref{sec:evaluation}, we show that sufficient rounds yield high quality subsets in practice, close to centralized greedy methods, despite not using centralized greedy on the final union of computed subsets, as done in prior methods. 
We lay out implementation details in~\cref{subsec:bounding-implementation}.

\noindent
\textbf{Computational complexity.}  Given a ground set of size \(|V|\), a budget \(k\), and \(k_g\) neighbors for which, for every point \(v_1\), \(s(v_1,.) > 0\), the complexity of the centralized algorithm is \(\mathcal{O}\left(|V| \log\left(|V|\right) + k \cdot k_g \cdot \log\left(|V|\right) \right)\) where the first term corresponds to the construction of priority queues and the second term corresponds to popping \(k\) points from the priority queue while adjusting \(k_g\) neighbors each time we pop a point.

For the distributed algorithm let \(m\) be the number of machines and \(r\) the number of rounds.
The computational complexity is then given by \(\mathcal{O}\left((r \cdot \left(|V|/m\right) \log\left(|V|/m\right) + r \cdot \frac{k}{m}\cdot k_g \cdot \log\left(|V|/m\right)\right)\).

\section{Implementing bounding and scoring}
\label{subsec:bounding-implementation}
An important aspect of our proposed algorithms is that we are able to support datasets where we \emph{cannot even hold the entire target subset in memory}.
In order to implement \textit{(i)} bounding and \textit{(ii)} subset scoring under this assumption, we use the Apache Beam programming model.
We assume that the number of neighbor interactions is limited, i.e., there is a (small) \(k \in \mathbb{N}\) s.t. for all \(v\in V\), \(|\left\{ v' \in V \mid s(v,v') > 0 \right\}| \leq k\).
We elaborate on this in~\Cref{sec:evaluation}.
For both bounding and scoring, the difficulty is that when iterating over the interacting neighbors of a data points, \emph{we cannot easily perform a \(\mathcal{O}(1)\) check whether the neighbor is in the subset}, as the subset is not in memory.
Instead, we need to cleverly use distributed joins.
We assume that we start with a \texttt{PCollection} of (node id, list of neighbors) tuples and a \texttt{PCollection} of (node id, utility) values.
A \texttt{PCollection} represents an immutable, conceptually infinitely-sized set of elements.
The set does not need to fit into DRAM, we can manipulate it using Beam operations without worrying about how the system processes the data.

\textbf{Bounding.}
We generate the \textit{fanned-out neighbor graph}, i.e., we iterate over the (node id, neighbor list) tuples, and for each neighbor in the list, emit a triple (neighbor id, node id, \(s(\text{node}, \text{neighbor})\)).
Note that the neighbor id becomes the triple key.
We keep track of a \texttt{PCollection} for the current partial result with (node, utility) tuples.
To get the minimum utility of all yet-unassigned data points, we need to find all neighbors of a point that are either in the partial solution (always considered) or not yet assigned.

To this end, we perform a distributed three-way join of the \texttt{PCollections} of the fanned neighbor graph, the current solution, and the currently unassigned points.
Within this join, we filter and invert the neighbor graph:
for a node \(a\), we know whether it is in the partial result by checking whether there is a join partner in that collection.
If not, we check if there is a join partner from the currently unassigned points.
If not, we discard the edge, since \(a\) is neither in the partial solution nor in the unassigned points, i.e., it has been removed in a shrink step.
Now, we iterate over all join partners from the neighbor graph, i.e., nodes \(b\) that were neighbors of \(a\).
We emit 4-tuples of the form (\(b\), \(a\), \(s(a,b)\), boolean indicating whether \(a\) is in the partial solution), recovering the original edges before generating the fanned-out neighbor graph, where \(a\) was a neighbor.
In these 4-tuples, \(a\) is guaranteed to be not discarded or in the partial solution, i.e., it is relevant for the minimum utility.
However, \(b\) might be already discarded or in the partial solution.
Hence, we perform another join of the 4-tuples with the unassigned points collection, and discard tuples where there is no join partner.
If there is a join partner, we can sample the neighborhood (when using approximate bounding), and use the boolean flag to know whether a neighbor should always be included to determine the minimum utility.
We then emit (node, min\_utility) tuples.
The maximum utility can be implemented analogously, but only needs to consider the points that are part of the partial solution.
The remaining parts of the bounding algorithm can be straightforwardly implemented using map operations.

\textbf{Scoring.} Distributed scoring is implemented similarly to the minimum utility calculation.
We generate the fanned-out neighbor graph, and then filter it by joining the solution, giving us all neighbors that are part of the solution.
We can then invert the result again, and reduce it to have a score per-datapoint, and then reduce this to an overall score by summing up the individual scores, as our function is decomposible.

\section{Evaluation}\label{sec:evaluation}
We demonstrate that our distributed algorithms achieve similar quality subsets as the centralized greedy algorithm.
We evaluate this by comparing them to objective obtained using the centralized algorithm. 
\neurips{We focus on studying the quality of the distributed algorithms compared to the centralized one, without training models, to limit the parameter space explored in our benchmarking (e.g., impact of embeddings, number of neighbors, hyperparametres such as \(\alpha, \beta\), on accuracy).}

\textbf{Datasets.} We use three datasets: (1) CIFAR-100~\cite{Krizhevsky2009CIFAR} with 100 classes and 50\,k points, (2) ImageNet~\cite{Deng2009ImageNet} with 1\,k classes and ca. 1.2\,M points, and (3) Perturbed-ImageNet with 1\,k classes and ca. 13\,B images. 
We obtain Perturbed-ImageNet by perturbing each point of ImageNet in embedding space into 10\,k vectors, leading to 13\,B embedding vectors for stress-testing our distributed algorithm.

\textbf{Embeddings, similarities, and utilities.}
For both CIFAR-100 and Imagenet, we generate predictions and embeddings for all points using a coarsely-trained ResNet-56~\cite{He2016DeepRL} trained on a random 10\,\% subset.
We employ SGD with Nesterov momentum 0.9, using 450/90 epochs for CIFAR/ImageNet.
The base learning rate is 1.0 for CIFAR and 0.1 for ImageNet, and is reduced by a tenth at epochs 15, 200, 300, 400 (CIFAR) and 5, 30, 69, 80 (ImageNet). We extract the penultimate layer features to generate $64$-dimensional embeddings for CIFAR, and $2048$-dimensional embeddings for ImageNet. 

We build a 10-nearest neighbor graph $(G, E)$ in the embedding space, using the fast similarity search by~\citet{Guo2020Search}. 
\neurips{Our experiments with a larger neighborhood \(k\) has not provided improvement in downstream real-world model training.
In any case, for billion-scale datasets, it is almost impossible to consider a fully connected graph for such massive collection of embeddings.
If we were to store 32-bit distance values for 1\,B datapoints, this would require 32\,exabytes of DRAM on each machine.
In contrast, storing only the 10-nearest neighbors requires only 40\,gigabytes of memory.
}
Note that this nearest neighbor graph is not symmetric. 
As \neurips{our distributed implementation of} bounding and scoring requires a symmetric graph (\Cref{subsec:bounding-implementation}), we symmetrize the graph, such that datapoints have a varying amount of, but at least 10 neighbors.
This leads to an average number of 15/16 neighbors for CIFAR-100/ImageNet, respectively.
Based on this graph, we set $s(v_1,v_2)$ in the submodular objective \(f(S) = \alpha \sum_{v \in S} u(v) - \beta \sum_{(v_1, v_2) \in E;v_1, v_2 \in S} s\left(v_1, v_2\right)\) to the cosine similarities of neighboring points.
We set \(\beta = 1 - \alpha\) and only mention the value of \(\alpha\) subsequently.

We use margin-based uncertainty~\cite{Scheffer2001} as the utility. In multi-class settings, margin provides high utility for points that are hard to classify, i.e., close to the decision boundaries: \(u(x_i) = 1 - \left(P\left(\text{top} \mid x_i\right) - P\left(\text{sec} \mid x_i\right)\right) \) where \(P\left(c \mid x_i\right)\) denotes the probability for class $c$ predicted by the model for the example \(x_i\), and \(top\) and \(sec\) denote the best and the second best classes as per the predictions.
Intuitively, a coarse model has already learned to classify easy points, while uncertain points are harder to learn and, therefore, more important.
\neurips{Note that the exact choice of similarity and utility scores (and number of neighbors) does not impact the comparison of the algorithm, as long as they are consistently used.}
We center the utilities by subtracting the minimum utility from all values.

\textbf{Normalization.} 
For each dataset, to compare the algorithms, for the same parameter group (dataset, \(\alpha/\beta\), and target subset size $k$), we map the objective from the centralized greedy to \(100\,\%\), and the lowest observed score to \(0\,\%\). This normalization enables us to uniformly interpret a percent point as a gain over the worst case, and emphasize instances where the basic centralized score is exceeded.

\subsection{Distributed Greedy Algorithm}\label{subsec:eval-greedy}
We fix the delta function to linear interpolation with a factor of \(0.75\), i.e., \(\Delta(|V|, r, \text{round}, k) = \ceil{0.75 \cdot (r - \text{round}) \cdot \frac{|V|-k}{r}}  + k\),
For an ablation, refer to~\Cref{app:linear-delta}.

\begin{figure*}[t]
    \centering
    \adjustbox{trim=0cm 0.2cm 0cm 0cm}{%
        \includesvg[width=0.80\textwidth]{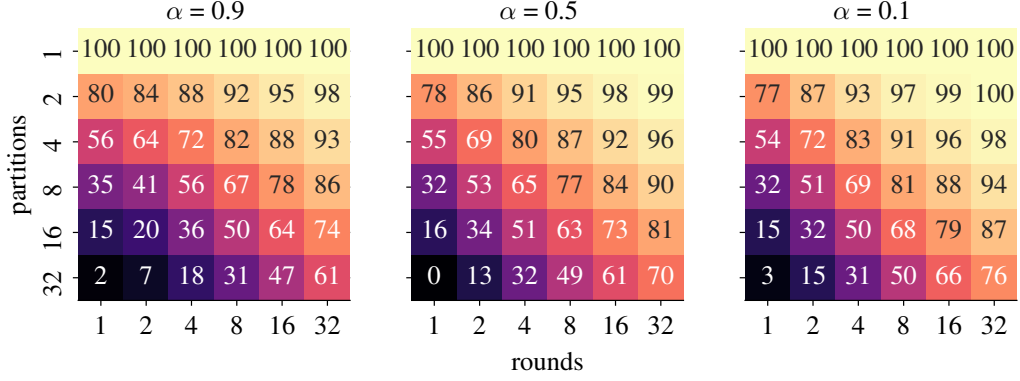}
    }
    \caption{{\it Normalized scores for finding a 10\,\% subset on CIFAR-100, depending on the number of partitions, rounds, and \(\alpha\). The full version can be found in~\Cref{fig:partition-rounds-cifar-full} (CIFAR) and~\Cref{fig:partition-rounds-imagenet-full} (ImageNet). Here, \( 100\) denotes the quality of centralized greedy algorithm.}} 
    \label{fig:parittions-rounds-cifar}
\end{figure*}

\textbf{Multiple partitions and multiple rounds.}
\Cref{fig:parittions-rounds-cifar} shows the normalized scores without adaptive partitioning depending on the subset size and \(\alpha\) resp. \(\beta\) on the CIFAR-100 dataset.
In all settings, \textit{(i)} fewer partitions lead to a higher score and \textit{(ii)} more rounds lead to a higher score. 
The reason is that when we partition the data into many sets, the neighborhood relation is lost across partitions. 
We refer to~\Cref{app:subset-viz} for a visualization.
Nevertheless, we still obtain high quality subsets by using multiple rounds.
For CIFAR, 2 partitions / 1 round have a 80\,\% score, but with 32 rounds, we obtain 98\,\% score.
Similarly, on ImageNet, the score increases from 86\,\% to 98\,\%.
We note that 32 partitions / 1 round does not have a score of 0 because the random partitioning in the adaptive partitioning experiment (same parameter group) leads to a slightly lower score.

\begin{figure*}[t]

    \centering
    \begin{adjustbox}{max width=0.80\textwidth,center}
    \adjustbox{trim=0cm 0.2cm 0cm 0cm}{%
        \includesvg[width=0.80\textwidth]{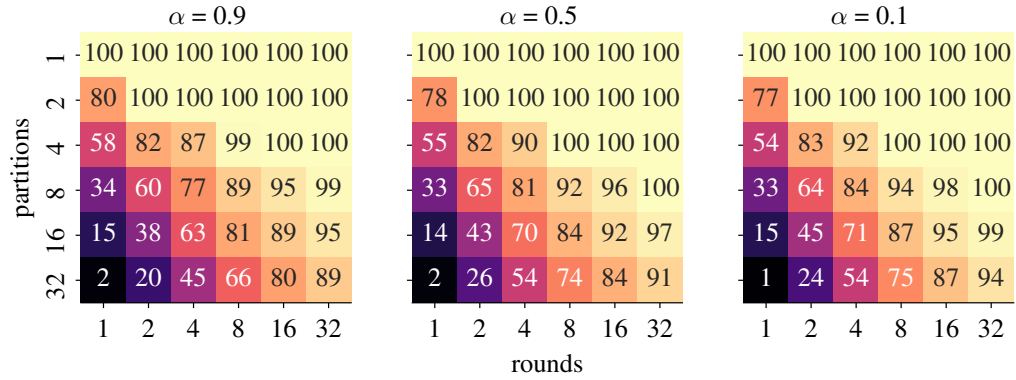}
    }
    \end{adjustbox}
    \caption{{\it Normalized scores for finding a 10\,\% subset on CIFAR-100, depending on the number of partitions, rounds, and \(\alpha\), using adaptive partitioning. The full version can be found in~\Cref{fig:adaptive-rounds-cifar-full} (CIFAR) and~\Cref{fig:adaptive-rounds-imagenet-full} (ImageNet).}}
    \vspace*{-0.3cm}
    \label{fig:adaptive-rounds-cifar}
\end{figure*}

For all settings, we find that the use of multiple rounds is more effective when the budget \(k\) is smaller.
For CIFAR and \(\alpha = 0.9\), when finding a 10\,\% subset, for 16 partitions, going from 1 to 32 rounds increases the score from 15\,\% to 74\,\%, while for a 50\,\% subset, it only increases the score from 8\,\% to 18\,\%.
For the other values of \(\alpha\), we find a similar trend.
For larger target sizes, the steps per round are smaller, such that there is less potential for improvement from repartitioning.
\neurips{Please refer to~\Cref{app:worst-case-partitioning} for a worst-case ablation showing the robustness of the algorithm.}

\emph{We observe that more partitions generally decrease and more rounds increase the submodular objective, especially for smaller subsets.}

\textbf{Adaptive partitioning.}
In \Cref{fig:adaptive-rounds-cifar}, we show that adaptive partitioning increases the score drastically since each round, we get closer to the centralized version and lose fewer edges in the neighbor graph. The benefit of adaptivity is higher the smaller the target subset size is, since for smaller subsets, fewer partitions are necessary. For example, for 10\,\% subsets, for both CIFAR and ImageNet, even with 32 partitions, we obtain a score of around 90\,\%, whereas without adaptive partitioning, the highest scores were 60 to 75\,\%. Furthermore, using adaptive partitioning is less resource-intensive, since it requires fewer parallel machines. 

\emph{Adaptive partitioning often reaches the centralized score, and uses fewer parallel resources.}

\vspace*{-0.3cm}
\subsection{Bounding}\label{subsec:eval-bounding}
\begin{table*}[t]
\centering
\caption{{\it Bounding results for \(\alpha = 0.9\). We achieve high quality results for most settings with uniform sampling, and occasionally even outperforming centralized methods.}}
\label{fig:bounding-tables}
\begin{adjustbox}{max width=0.9\textwidth,center}

\begin{tabular}{l|l|rrr||rrr}
                                                                        &                                                                                                                                            & \multicolumn{3}{c||}{CIFAR100}                                                                                                                                                                                                                                                                                                                                                                                       & \multicolumn{3}{c}{ImageNet}                                                                                                                                                                                                                                                                                                                                                                                                      \\ 
\hline
\rowcolor[rgb]{0.878,0.878,0.878} \multicolumn{1}{c|}{Type of Sampling} &                                                                                                                                            & \multicolumn{1}{c}{10\,\% Subset}                                                                                                                                                  & \multicolumn{1}{c}{50\,\% Subset}                                                                                 & \multicolumn{1}{c||}{80\,\% Subset}                                                                         & \multicolumn{1}{c}{10\,\% Subset}                                                                               & \multicolumn{1}{c}{50\,\% Subset}                                                                                    & \multicolumn{1}{c}{80\,\% Subset}                                                                                                                                                        \\ 
\hline
No sampling                                                             & \begin{tabular}[c]{@{}l@{}}Included / Excluded Points\\\# of Grow / Shrink Rounds\\Score\end{tabular}                                      & \begin{tabular}[c]{@{}r@{}}0 / 10\,769\\1 / 16\\100.01\,\%\end{tabular}                                                                                                            & \begin{tabular}[c]{@{}r@{}}0 / 0\\~1 / 1\\100.0\,\%\end{tabular}                                                  & \begin{tabular}[c]{@{}r@{}}2\,002 / 0\\9 / 2\\100.55\,\%\end{tabular}                                       & \begin{tabular}[c]{@{}r@{}}0 / 0\\~1 / 1\\100.0\,\%\end{tabular}                                                & \begin{tabular}[c]{@{}r@{}}0 / 0~\\1 / 1\\100.0\,\%\end{tabular}                                                     & \begin{tabular}[c]{@{}r@{}}8043 / 0\\5 / 2\\100.1\,\%\end{tabular}                                                                                                                       \\
\rowcolor[rgb]{0.878,0.878,0.878} 30\,\% Uniform                        & \begin{tabular}[c]{@{}>{\cellcolor[rgb]{0.878,0.878,0.878}}l@{}}Included / Excluded Points\\\# of Grow / Shrink Rounds\\Score\end{tabular} & \begin{tabular}[c]{@{}>{\cellcolor[rgb]{0.878,0.878,0.878}}r@{}}3 / 25\,743\\2 / 10\\100.0\,\%\end{tabular}                                                                        & \begin{tabular}[c]{@{}>{\cellcolor[rgb]{0.878,0.878,0.878}}r@{}}24\,972 / 14\,896\\9 / 7\\~97.39\,\%\end{tabular} & \begin{tabular}[c]{@{}>{\cellcolor[rgb]{0.878,0.878,0.878}}r@{}}39\,999 / 0\\~4 / 2\\85.95\,\%\end{tabular} & \begin{tabular}[c]{@{}>{\cellcolor[rgb]{0.878,0.878,0.878}}r@{}}16 / 756\,625\\3 / 10\\100.01\,\%\end{tabular}  & \begin{tabular}[c]{@{}>{\cellcolor[rgb]{0.878,0.878,0.878}}r@{}}633\,027 / 274\,635\\34 / 12\\99.77\,\%\end{tabular} & \begin{tabular}[c]{@{}>{\cellcolor[rgb]{0.878,0.878,0.878}}r@{}}1\,024\,932 / 0\\~5 / 2\\92.72\,\%\end{tabular}                                                                          \\
70\,\% Uniform                                                          & \begin{tabular}[c]{@{}l@{}}Included / Excluded Points\\\# of Grow / Shrink Rounds\\Score\end{tabular}                                      & \begin{tabular}[c]{@{}r@{}}\textcolor[rgb]{0.129,0.129,0.129}{0 / 17\,785}\\\textcolor[rgb]{0.129,0.129,0.129}{1 / 9}\\\textcolor[rgb]{0.129,0.129,0.129}{100.06\,\%}\end{tabular} & \begin{tabular}[c]{@{}r@{}}0 / 0\\1 / 1\\~100.0\,\%\end{tabular}                                                  & \begin{tabular}[c]{@{}r@{}}30\,724 / 0\\~165 / 2\\~103.51\,\%\end{tabular}                                  & \begin{tabular}[c]{@{}r@{}}0 / 470\,133\\~1 / 14\\~100.0\,\%\end{tabular}                                       & \begin{tabular}[c]{@{}r@{}}0 / 0\\1 / 1\\~100.0\,\%\end{tabular}                                                     & \begin{tabular}[c]{@{}r@{}}192\,019 / 0\\~189 / 2\\102.13\,\%\end{tabular}                                                                                                               \\
\rowcolor[rgb]{0.878,0.878,0.878} 30\,\% Weighted                       & \begin{tabular}[c]{@{}>{\cellcolor[rgb]{0.878,0.878,0.878}}l@{}}Included / Excluded Points\\\# of Grow / Shrink Rounds\\Score\end{tabular} & \begin{tabular}[c]{@{}>{\cellcolor[rgb]{0.878,0.878,0.878}}r@{}}3 / 25\,729\\~2 / 8\\100.0\,\%\end{tabular}                                                                        & \begin{tabular}[c]{@{}>{\cellcolor[rgb]{0.878,0.878,0.878}}r@{}}24\,996 / 14\,853\\~6 / 9\\75.59\,\%\end{tabular} & \begin{tabular}[c]{@{}>{\cellcolor[rgb]{0.878,0.878,0.878}}r@{}}39\,999 / 0\\4 / 2\\81.79\,\%\end{tabular}  & \begin{tabular}[c]{@{}>{\cellcolor[rgb]{0.878,0.878,0.878}}r@{}}16 / 756\,511\\~2 / 10\\100.01\,\%\end{tabular} & \begin{tabular}[c]{@{}>{\cellcolor[rgb]{0.878,0.878,0.878}}r@{}}639\,492 / 273\,628\\9 / 11\\81.95\,\%\end{tabular}  & \begin{tabular}[c]{@{}>{\cellcolor[rgb]{0.878,0.878,0.878}}r@{}}1\,024\,934 / 0\\4 / 1\\~78.52\,\%\end{tabular}                                                                          \\
70\,\% Weighted                                                         & \begin{tabular}[c]{@{}l@{}}Included / Excluded Points\\\# of Grow / Shrink Rounds\\Score\end{tabular}                                      & \begin{tabular}[c]{@{}r@{}}0 / 17\,748\\1 / 9\\100.06\,\%\end{tabular}                                                                                                             & \begin{tabular}[c]{@{}r@{}}0 / 0\\1 / 1\\100.0\,\%\end{tabular}                                                   & \begin{tabular}[c]{@{}r@{}}40\,000 / 0\\6 / 1\\~88.17\,\%\end{tabular}                                      & \begin{tabular}[c]{@{}r@{}}0 / 469\,860\\1 / 14\\~100.0\,\%\end{tabular}                                        & \begin{tabular}[c]{@{}r@{}}0 / 0\\1 / 1\\100.0\,\%\end{tabular}                                                      & \begin{tabular}[c]{@{}r@{}}\textcolor[rgb]{0.129,0.129,0.129}{1\,024\,904 / 0}\\\textcolor[rgb]{0.129,0.129,0.129}{12 / 2}\\\textcolor[rgb]{0.129,0.129,0.129}{~95.73\,\%}\end{tabular} 
\end{tabular}

\end{adjustbox}
\end{table*}

\Cref{fig:bounding-tables} shows the results for \(\alpha=0.9\).
To evaluate each configuration, we investigate the number of included/excluded datapoints, and the centralized score (1 partition/1 round) when processing the bounding result with the greedy algorithm.
For other configurations, see~\Cref{app:figures-bounding}.

\textbf{Exact bounding.}
For CIFAR (50\,k points), in the 10\,\% subset case, bounding excludes ca. 22\,\% of the points in 16 rounds.
In the 50\,\% subset case, it does not make any decision, and for the 80\,\% subset, it includes 4\,\% in 9 rounds. 
For ImageNet (1.2\,M points), it does not make a decision for 10\,\% and 50\,\% subsets, and includes less than 1\,\% of the dataset in 5 rounds.
For both datasets, in the 80\,\% subset case, the final score is slightly higher than the centralized non-bounding score, just by including a few data points.

Overall, exact bounding only includes or excludes a few points for very small or very large subsets, respectively. We observe this behavior because a smaller target size leads to easier requirements for shrinking, which results in more excluded points (\Cref{alg:grow-bounding}). The same argument holds for large target sizes and inclusion. Intuitively, when the target subset size is extreme, it is easier for the algorithm to make a decision, since overall more points are included/excluded.

\textbf{Approximate bounding.}
We test approximate bounding with uniform sampling, i.e., all neighbors of a data point have the same chance of being sampled, and weighted sampling, i.e., the sampling probability is uniform to the pairwise interaction between the neighbors.
For both sampling types, we test sampling a 30\,\% and 70\,\% neighborhood.
Generally, for both CIFAR and ImageNet, the 30\,\% neighborhood is able to include and exclude many points.
However, the 70\,\% neighborhood still struggles to make decisions, especially for the 50\,\% subset setting.
We also find that the larger the neighborhood, i.e., the more information we use, the higher the score is.

In the 10\,\% subset setting, for both CIFAR and ImageNet, when sampling a 30\,\% neighboorhood, the algorithm excludes around 50\,\% of the dataset (22\,\% and \(<\) 1\,\% for exact bounding on CIFAR and ImageNet), reducing the load on the centralized algorithm and requiring fewer partitions at the start.
In the 80\,\% subset setting, the algorithm often finds (almost) the entire subset without the need for the greedy algorithm. 
Interestingly, it struggles to do so with 70\,\% uniform sampling, where for ImageNet, it includes ten times fewer points and takes 189 rounds, compared to five to twelve rounds in the other settings. 
Weighted sampling helps the algorithm to converge faster, but sometimes has a worse score than uniform sampling, which we attribute to the bias in the sampling strategy that not always is optimal.
Last, for 30\,\% neighborhoods, the algorithm is even able to both include and exclude data points in the 50\,\% subset setting, which is something that exact bounding and even the 70\,\% neighborhood struggle with.
Approximate sampling empirically performs well, keeping scores of over 90\,\%, and is implementable in a parallel setting.

In summary, approximate bounding includes and excludes many more points than exact bounding at the cost of a sometimes slightly lower objective. 
For large subsets, bounding often finds the entire subset on its own. 
However, as it is massively parallelizable, it still performs well from a runtime-objective tradeoff standpoint.

\textbf{Lowering the utility coefficient \(\alpha\).}
For \(\alpha \in \left\{ 0.1, 0.5 \right\}\) we find that no configuration includes/excludes any points. The smaller values of \(\alpha\) increase the role of the pairwise terms. This makes the minimum and maximum utilities drift apart, leading to less growing/shrinking of points.

\emph{Bounding often performs better, and occasionally even better than the centralized method, when the utility dominates the objective function.}

\subsection{Dataset with 13 Billion Points}\label{subsec:huge-dataset}
We demonstrate the scalability of our approach on a dataset with around 13 billion datapoints.
For the distributed submodular algorithm, we use 16 partitions with 350\,GB of memory per partition. 
We test 1, 2, and 8 rounds for \(\alpha = 0.9\).
We cannot normalize the scores since we are unable to determine the centralized score.
See~\Cref{app:runtime} for a runtime discussion.

\textbf{10\,\% subset.}
The score increases from 1\,058\,841\,312 (1 round) to 1\,092\,474\,410 (2 rounds), up to 1\,145\,682\,717 (8 rounds).
We also test the bounding procedure. Exact bounding includes 0.007\,\% of the 13 billion points and excludes 10\,\%.
Approximate bounding with a 30\,\% neighborhood includes 0.7\,\% of the points and excludes 60\,\% for both uniform and weighted sampling. All bounding approaches reach a score of slightly above 100\,\% of the score without bounding after 8 rounds.

\textbf{50\,\% subset.} 
For the distributed greedy algorithm, the score increases from 4\,168\,989\,874 (1 round), to 4\,200\,071\,672 (2 rounds), up to 4\,250\,047\,523 (8 rounds).

\begin{table*}[t]
\centering
\caption{{\it Normalized (non adaptive/adaptive) scores for random partitioning and worst-case partitioning in the first round on CIFAR-100.}}
\small
\label{tab:worstcase-app}
\begin{tabular}{l|rrrr}
\multicolumn{1}{c|}{}     & \multicolumn{1}{c}{\textbf{1 round}} & \multicolumn{1}{c}{\textbf{8 rounds}} & \multicolumn{1}{c}{\textbf{16 rounds}} & \multicolumn{1}{c}{\textbf{32 rounds}}  \\ 
\hline
random partitioning       & 27\,\% / 27\,\%                      & 63\,\% / 89\,\%                       & 74\,\% / 94\,\%                        & 83\,\% / 97\,\%                         \\
solution in one partition & 10\,\% / 10\,\%                      & 60\,\% / 87\,\%                       & 71\,\% / 91\,\%                        & 81\,\% / 94\,\%                         
\end{tabular}
\vspace{-0.5cm}
\end{table*}

\subsection{Worst-Case Partitioning Ablation}\label{app:worst-case-partitioning}

We give further empirical evidence that the distributed greedy algorithm is resilient against bad partitioning, i.e., assigning all optimal points into the same partition.
Intuitively, having the optimal solution in a single partition seems unlikely, especially at the scale of multiple billions of data points, and multiple rounds protect against that. 
Even in the unlikely case of a bad assignment in the first round, the probability of that reoccurring in subsequent rounds is very low.
Deriving a theoretical bound is an interesting future work direction.

We run the following experiment.
On CIFAR-100, we select the centralized greedy 10\,\% subset and assume this is the best solution, since we always use that as the reference point.
Then, we run the distributed greedy algorithm, where one partition is the centralized solution. 
This means we have 10 partitions, and one of them contains the centralized solution and the remaining 9 partitions are chosen randomly.
In subsequent rounds, the dataset is then repartitioned randomly.
This simulates a worst-case initial random assignment, since it minimizes the number of data points we can pick from the centralized (reference) solution.

The results can be found in~\Cref{tab:worstcase-app}.
We give both the non adaptive / adaptive partition score, in the same normalization scheme as in the previous subsections.
For the single round case, there is a 17\,percent point score difference in the worst case. 
When using multiple rounds, the penalty for this worst-case partitioning is only 2-3\,percent points. 
This shows the robustness of the multi-round approach.

\section{Discussion}\label{sec:conclusion}
We present a highly-parallelizable and distributed bounding algorithm for optimizing pairwise submodular functions for subset selection, which does not assume the subset fits in the main memory of a single machine.
We also propose a multi-round, partitioning-based, distributed algorithm that yields high-quality subsets with minimal to no quality loss compared to centralized algorithms. 
We conduct a comprehensive analysis of both the bounding and the distributed greedy algorithm and providing insights on achieving high quality subsets. %
We find that using more rounds help to reach near-centralized scores, especially when adaptively adjusting the number of parallel partitions. In some cases, approximate bounding is able to exclude over 50\,\% of the dataset, and sometimes solves the entire problem.

Despite proving guarantees for the bounding algorithm, showing approximation guarantees for the distributed greedy algorithm remains future work.
Occasionally, we observe that the approximate bounding can even outperform centralized methods, and we plan to explore this further.

\FloatBarrier

\section*{Acknowledgments}
We would like to thank Sameer Agarwal, Sanjiv Kumar, Christian Tjandraatmadja, Ayan Chakrabarti, Daniel Glasner, and Morteza Zadimoghaddam for their valuable help.
We thank the reviewers for their helpful feedback.
Maximilian Böther was partially supported by the Swiss National Science Foundation (project number 200021\_204620).

\bibliography{main}
\bibliographystyle{mlsys2025}

\newpage
\appendix

\section{Monotonicity of the objective function}
\label{lb.monotonic}
For any subset $S$ of the ground set $V$, the objective value at $S$ is given as
\begin{align}
    f(S) = \alpha \sum_{v \in S} u(v) - \beta \sum_{(v_1, v_2) \in E;v_1, v_2 \in S} s\left(v_1, v_2\right).
\end{align}

In many downstream applications, we choose $\alpha = 0.9$ and $\beta = 0.1$. For such large $\alpha$ and small $\beta$, the unary terms will dominate the pairwise terms, and the function will be monotonic. In cases where the above function is not monotonic, we can  add a constant offset to all the unary terms to make it monotonic. For example, we can consider the following offset:

\begin{equation}
   \delta  = \frac{\beta}{\alpha} \max_{l\in V} \sum_{j:(l,j) \in E}s(l,j)
\end{equation}

By adding this constant offset to all the nodes, we can ensure that the adjusted function is both submodular and monotonic:
\begin{align}
    f(S) = \alpha \sum_{v \in S} (u(v)+\delta) - \beta \sum_{(v_1, v_2) \in E;v_1, v_2 \in S} s\left(v_1, v_2\right).
\end{align}

This ensures that for $B \subseteq A$, $f(B) \le f(A)$.
In this case, the standard approximation guarantee of $f(S)\ge (1-\frac{1}{e})f(S_{\text{OPT}})$ becomes $f(S)+k\delta \ge (1-\frac{1}{e})(f(S_{\text{OPT}} + k\delta)$.

\section{Theoretical Analysis of Approximate Bounding}
\label{thm.proof}

In this section, we provide the proof for~\Cref{thm:main}.

\begin{proof}
We first bound the performance of the algorithm with respect to a generic approximate algorithm for computing $U_{\text{min}}(v)$ and then instantiate the selection procedure to be random i.i.d. sampling to get the desired final bound. 

For any set $S$, the objective value at $S$ is
\begin{align}
    f(S) = \alpha \sum_{v \in S} u(v) - \beta \sum_{(v_1, v_2) \in E;v_1, v_2 \in S} s\left(v_1, v_2\right).
\end{align}
Alternately, we can write the above as
\begin{align}
    f(S) = \sum_{v \in S} \tilde{u}(v,S),
\end{align}
where $\tilde{u}(v,S)$ is defined as
\begin{align}
    \tilde{u}(v,S) = \alpha u(v) - \frac{\beta}{2} \sum_{v_2 \in S} s(v, v_2).
\end{align}

Next consider a generic procedure for computing $U_{\text{min}}$ values that outputs values $\tilde{U}_{\text{min}}(v)$ such that $U_{\text{min}}(v) \leq \tilde{U}_{\text{min}}(v) \leq r U_{\text{min}}(v)$ holds for all $v \in V$, for some $r > 1$. 
Next consider a surrogate objective $\hat{f}(S)$ defined as
\begin{align}
    \hat{f}(S) &= \sum_{v \in S} \hat{u}(v,S)\\
    &=\vcentcolon \sum_{v \in S} \max \big(\tilde{u}(v,S), \alpha \tilde{U}_{\text{min}}(v) \big).
\end{align}

Notice that the surrogate objective has the property that for any set $S$, $f(S) \leq \hat{f}(S) \leq r f(S)$. Furthermore, with respect to the surrogate objective the grow and reduce operations are performing exact bounding as no element of the optimal solution will be discarded. As a result, the algorithm will output a set $S$ such that 
\begin{align}
    f(S) \geq \frac{1}{2r}f(S^*).
\end{align}

Finally, it remains to bound the value of $r$.
We will show that, with high probability, $r \leq (1 + \gamma(1-p)^2)$.
The term $k_g$ in the exponent refers to the minimum degree of the graph.
To show this bound, we use the fact that we sample the neighbors uniformly at random with probability $p$. 
Consider a particular vertex $v$ and let $\mu(v) = \sum_{v_1 \in N(v)} s(v,v_1)$. 
In the approximate bounding algorithm we sample each neighbor of a vertex independently with probability $p$.
For any $v_1 \in N(v)$, let $y_{v_1}$ be a $\{0,1\}$ valued random variable such that if $y_{v_1}=1$ then it implies that $v_1$ was sampled for the vertex $v$.
Notice that $\mathbb{P}(y_{v_1}=1) = p$ and that these are independent random variables.

Next notice that in the algorithm when we compute the min and max utilities of any vertex we use the sampled neighbors. Hence we have that for vertex $v$
\begin{align}
    \tilde{U}_{\min} = u(v) - \frac{\beta}{\alpha}\sum_{v_1 \in N(V)} y_{v_1} s(v,v_1).
\end{align}

In order to bound the value of $r$ we need to bound the ratio of $\tilde{U}_{\min}$ to $U_{\min}$, i.e.,
\begin{align}
    \frac{\tilde{U}_{\min}}{U_{\min}} &= \frac{u(v) - \frac{\beta}{\alpha}\sum_{v_1 \in N(V)} y_{v_1} s(v,v_1)}{u(v) - \frac{\beta}{\alpha}\sum_{v_1 \in N(V)} s(v,v_1)}
\end{align}

Since the numerator above is a random quantity we can only hope for an upper bound that holds with high probability (over the sampling of the neighbors) and for that we need to use tail bounds.
In particular recall the following Chernoff/Hoeffding's inequality

\begin{lemma}
Let $X_1, X_2, \ldots X_n$ be independent real valued random variables taking values in $[a,b]$. Let $X = \sum_i X_i$ and let $\mu = \mathbb{E}[X]$. Then for any $t > 0$,
\begin{align*}
    \mathbb{P}[X \leq \mu - t] \leq e^{-2\frac{t^2}{n(b-a)^2}}.
\end{align*}
\end{lemma}

We will apply the above inequality to the sum $X = X_1 + X_2 + \ldots X_n$ where $X_i = y_i s(v,v_i)$ for $v_i \in N(v)$. Furthermore, we will denote the full sum by $S$, i.e., $S = \sum_{v_1 \in v} s(v,v_1)$.

In the above notation we get that in order to bound $r$ we need to bound
\begin{align}
    \frac{\tilde{U}_{\min}}{U_{\min}} &= \frac{u(v) - \frac{\beta}{\alpha}X}{u(v) - \frac{\beta}{\alpha}S}.
\end{align}

Note that each random variable takes values in $[a,b]$ because of our assumption about the boundedness of the similarities.
Next we have that $\mu = \sum_i \mathbb{E}[y_i]s(v,v_i) = p S$.

By the above mentioned tail bound we have
\begin{align}
    \mathbb{P}[X \leq \mu - t] &\leq e^{-2\frac{t^2}{|N(v)| (b-a)^2}}.
\end{align}

Let's set $t = \delta p S$ where we will set $\delta > 0$ later in the proof. Then we get that
\begin{align}
    \mathbb{P}[X \leq p (1-\delta) S] &\leq e^{-2\frac{\delta^2 p^2 S^2}{|N(v)| (b-a)^2}}.
\end{align}

Next, note that because of the fact that the min degree in the graph is $k_g$ and the minimum similarity values are at least $a$ we have $S \geq a |N(v)|$ and that $|N(v)| \geq k_g$. Substituting this above we get that
\begin{align}
    \mathbb{P}[X \leq p (1-\delta) S] &\leq e^{-2\frac{\delta^2 p^2 a^2 k_g}{(b-a)^2}}.
\end{align}

The above analysis was for a single vertex. Setting $\delta=1-p$ and via a union bound over all the $|V|$ vertices we get that with probability at least $1 - |V| e^{-2 \frac{(1-p)^2 p^2 a^2 k_g}{(b-a)^2}}$, for every vertex $v$ the corresponding $X$ variable is at least $p(1-\delta)S = p^2 S$ in value.

Finally, conditioned on the above event we are ready to bound $r$. We have
\begin{align}
    \frac{\tilde{U}_{\min}}{U_{\min}} &= \frac{u(v) - \frac{\beta}{\alpha}X}{u(v) - \frac{\beta}{\alpha}S}\\
    &\leq \frac{u(v) - p^2\frac{\beta}{\alpha}S}{u(v) - \frac{\beta}{\alpha}S}.
\end{align}

Next define $x = \frac{\beta S}{\alpha u(v)}$. Then the above can be rewritten as
\begin{align}
    \frac{\tilde{U}_{\min}}{U_{\min}} &\leq \frac{1 - p^2 x}{1-x}\\
    &= 1 + \frac{x - p^2 x}{1-x}\\
    &= 1 + \frac{x(1-p^2)}{1-x} \label{eq:proof-final}.
\end{align}

Next, recall our assumption that the ratio of the max and the min utilities is bounded by $\gamma$. This means that
\begin{align}
    \frac{u(v)}{u(v) - \frac{\beta}{\alpha}S} \leq \gamma.
\end{align}

In terms of $x$ this implies that $x \leq 1 - \frac{1}{\gamma}$, or in other words $\frac{x}{1-x} \leq \gamma$.

Substituting back into Eq.~\ref{eq:proof-final} we get that
\begin{align}
r = \frac{\tilde{U}_{\min}}{U_{\min}} &\leq 1 + \gamma (1-p^2). 
\end{align}

\end{proof}

\clearpage
\newpage

\section{Subset Visualization}\label{app:subset-viz}

\begin{figure*}[ht]
    \centering
    \includegraphics[width=0.7\textwidth]{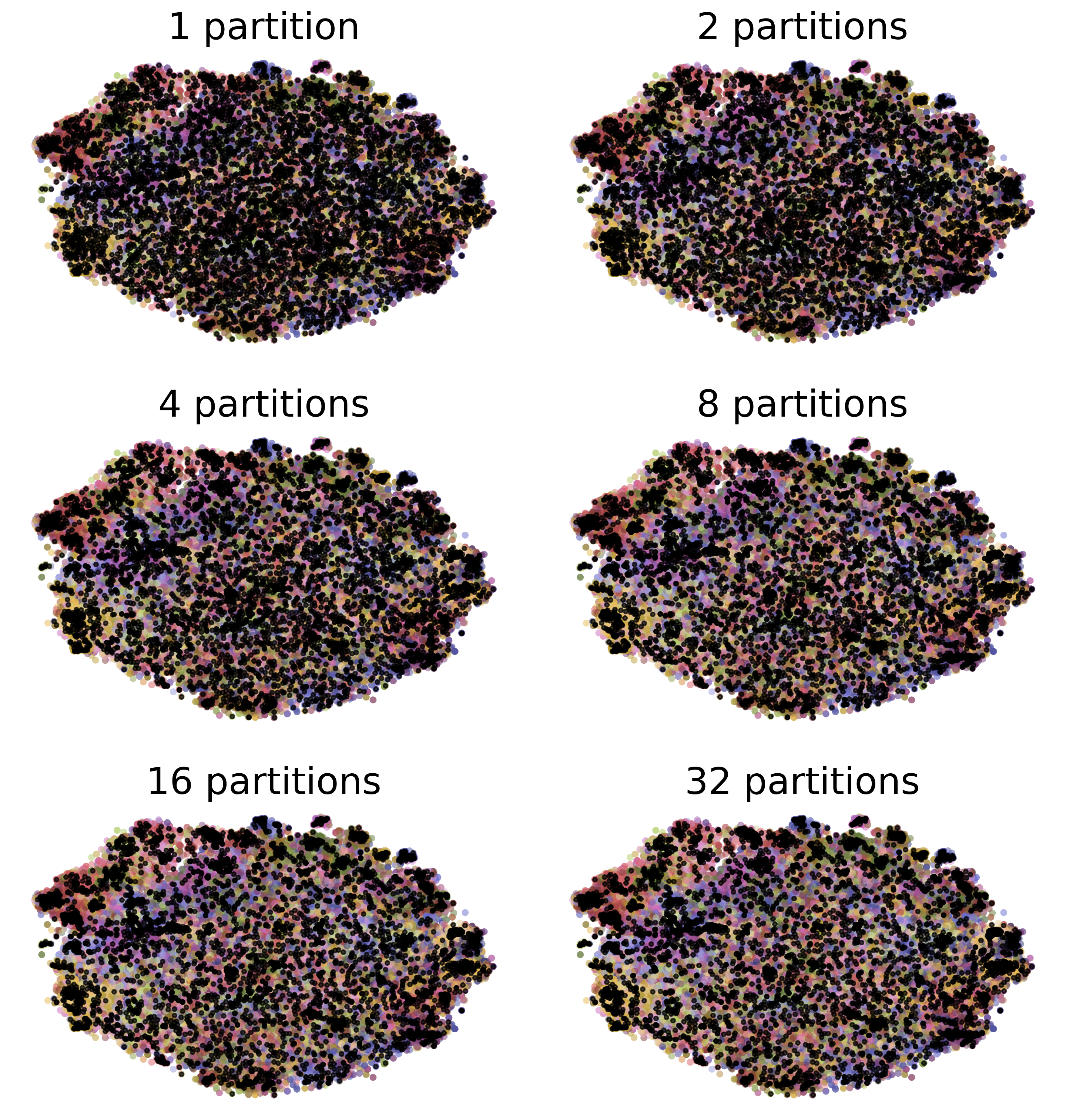}
    \caption{A rasterized visualization of the chosen 5\,000 points out of the 50\,000 points in CIFAR-100. The points are colored by label, and chosen data points are depicted as black.} 
    \label{fig:subset-10-09-embeddings}
\end{figure*}

In this section, we visualize the chosen subset on the example of finding a 10\,\% subset of CIFAR-100 with \(\alpha = 0.9\), depending on the number of partitions for one round.
To this end, we take the embeddings of the data points and reduce them to the 2-dimensional plane using t-SNE.
We use scikit-learn's implementation with default settings, i.e., PCA initialization and automatic learning rate.
The results are shown in~\Cref{fig:subset-10-09-embeddings}.
The centralized version chooses data points more uniformly across the plane, while using more partitions creates local clusters.
This is because the random partitioning loses information about local edges.
The algorithm per partition focuses more on the utility of data points, since it cannot reason about their influence on the diversity score.
Overall, this leads to local utility clusters in case of many partitions, whereas the single partition algorithm would have distributed the points more evenly across the plane.

\clearpage
\newpage

\section{Runtime Analysis}\label{app:runtime}

\begin{table*}
\centering
\caption{Runtimes for internal distributed implementation for different algorithms on the 13\,B dataset}
\label{tab:runtimes}
\begin{adjustbox}{max width=\textwidth}
\begin{tabular}{l|r|r}
\multicolumn{1}{c|}{Algorithm}                                   & \multicolumn{1}{c|}{10\% Subset Runtime~} & \multicolumn{1}{c}{50\% Subset~Runtime}  \\ 
\hline
Approximate bounding with uniform sampling                       & 19.61\,h                                  & -                                        \\
Approximate bounding with weighted sampling                      & 21.31\,h                                  & -                                        \\
8 rounds of distributed greedy algorithm after uniform bounding  & 33.46\,h                                  & -                                        \\
8 rounds of distributed greedy algorithm after~weighted~bounding & 27.2\,h                                   & -                                        \\
8 rounds of distributed greedy algorithm without bounding        & 40.72\,h                                  & 48.22\,h                                 \\
2 rounds of distributed greedy algorithm without bounding        & 20.45\,h                                  & 16.32\,h                                 \\
1 rounds of distributed greedy algorithm without bounding        & 9.86\,h                                   & 12.7\,h                                 
\end{tabular}
\end{adjustbox}
\end{table*}

\neurips{
In~\Cref{tab:runtimes}, we give the runtimes of our internal implementation for the large dataset (13\,B) experiments (\Cref{subsec:huge-dataset}).
Since we run the distributed algorithms in our large internal heterogeneous clusters, we cannot conduct an accurate runtime analysis due to scheduling issues with many nodes and the scale of the benchmarking matrices.
We can only state the runtime of a single run.
The algorithm finishes within roughly 2 days for most settings, even on the 13 billion dataset.
For the small datasets such as CIFAR and ImageNet, the experiments can be completed in a few seconds. %
}

\clearpage
\newpage

\section{Ablation on \(\Delta\)}\label{app:linear-delta}
In this section, we perform an ablation study on the delta function (\Cref{subsec:distri-algo}).
We  use the linear delta function from~\Cref{sec:evaluation}, but vary the interpolation factor which we set to \(\gamma = 0.75\) in the evaluation.
This means for \(\gamma \in \{0.25, 0.5, 1\}\), we test \(\Delta(|V|, r, r_{\text{curr}}, k) = \ceil{\gamma \cdot (r - r_{\text{curr}}) \cdot \frac{|V|-k}{r}}  + k\).
Note that other choices of \(\Delta\) are possible, but we limit ourselves to linear interpolation.
We only evaluate the influence of \(\gamma\) on 10\,\% and 50\,\% subsets since the influence of the delta function for large subsets is limited.

\begin{figure*}[ht]
    \centering
    \includesvg[width=0.8\textwidth]{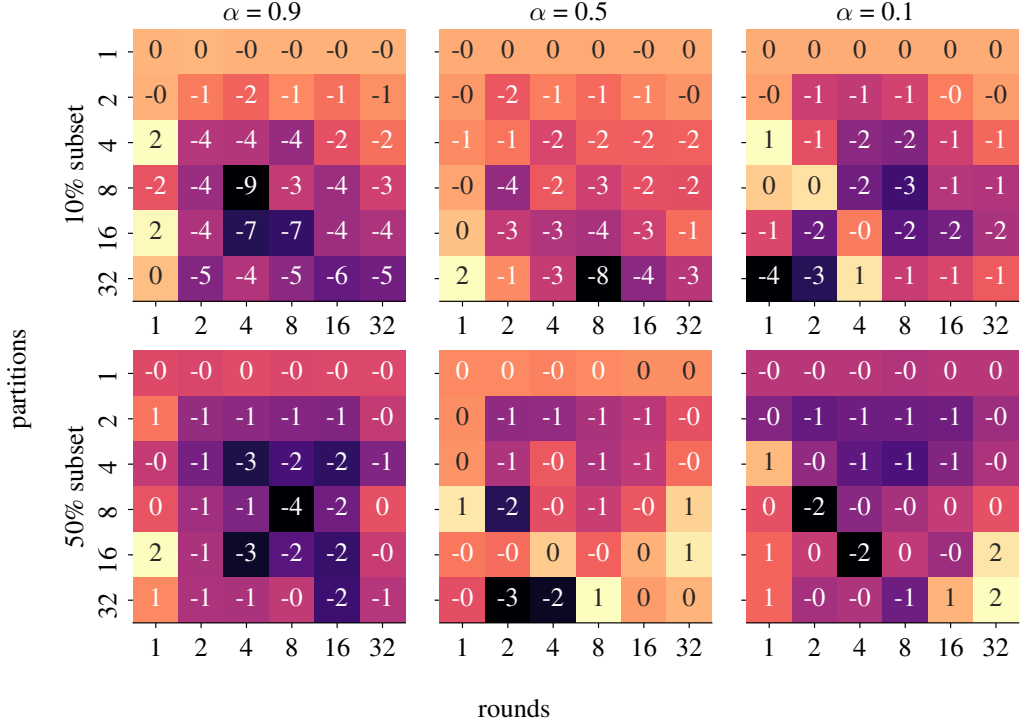}
    \caption{Difference in normalized score of \(\gamma = 1\) to \(\gamma = 0.75\) on CIFAR-100, depending on the subset size, the number of partitions, rounds, and \(\alpha\).} 
    \label{fig:cifar-partition-rounds-noadap-gamma1}
\end{figure*}

\begin{figure*}[ht]
    \centering
    \includesvg[width=0.8\textwidth]{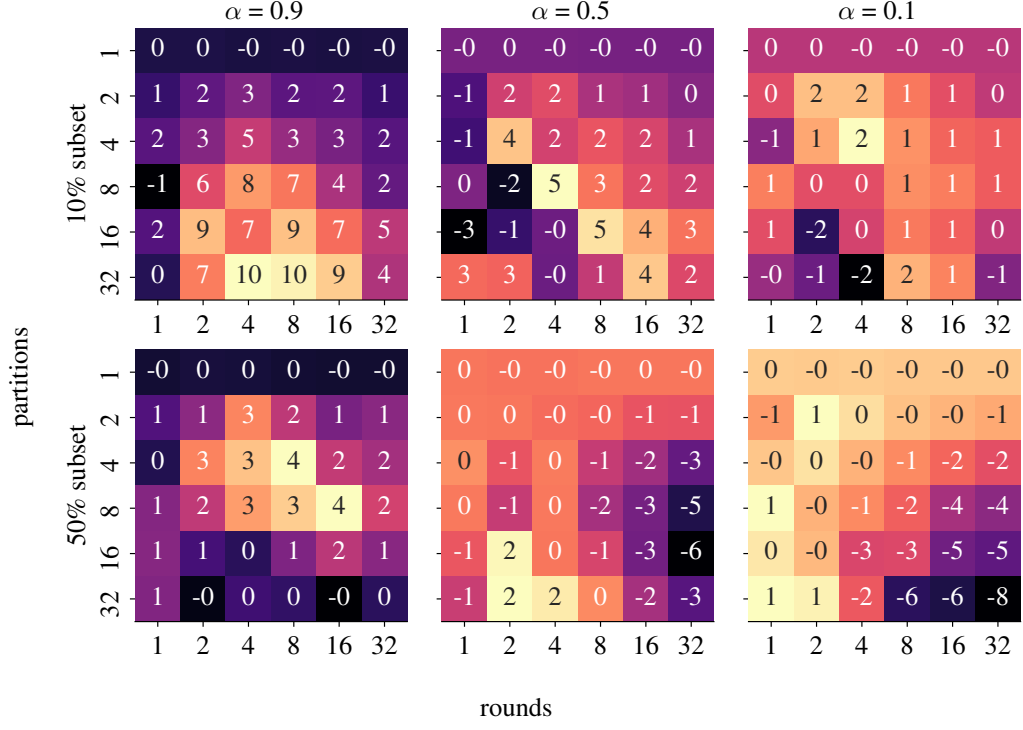}
    \caption{Difference in normalized score of \(\gamma = 0.5\) to \(\gamma = 0.75\) on CIFAR-100, depending on the subset size, the number of partitions, rounds, and \(\alpha\).} 
    \label{fig:cifar-partition-rounds-noadap-gamma5}
\end{figure*}

\begin{figure*}[ht]
    \centering
    \includesvg[width=0.8\textwidth]{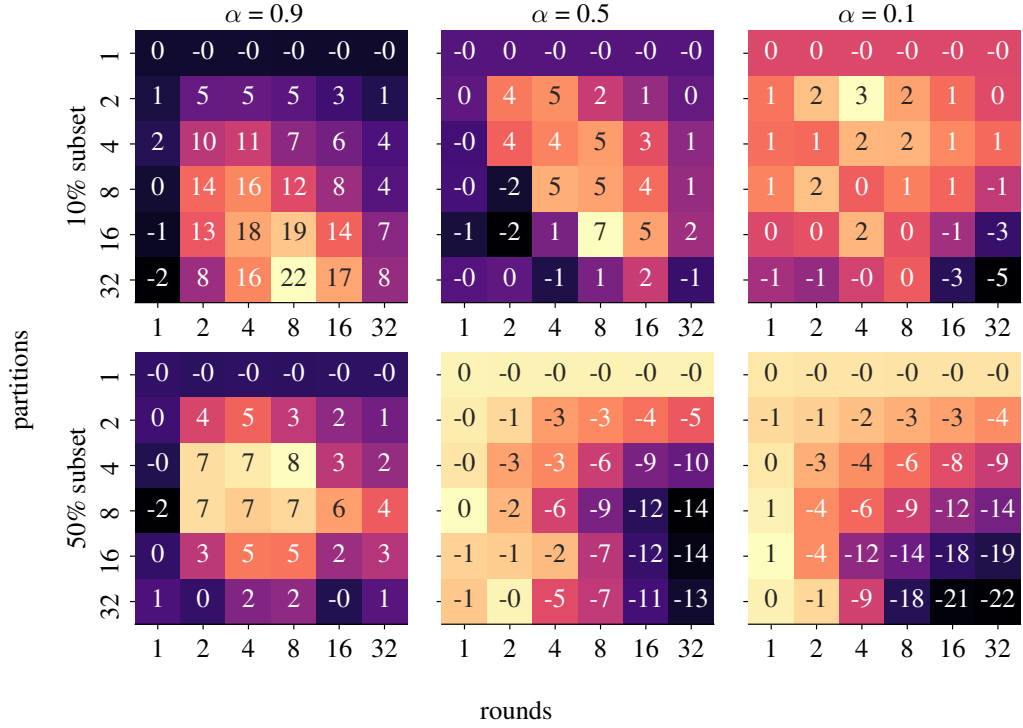}
    \caption{Difference in normalized score of \(\gamma = 0.25\) to \(\gamma = 0.75\) on CIFAR-100, depending on the subset size, the number of partitions, rounds, and \(\alpha\).} 
    \label{fig:cifar-partition-rounds-noadap-gamma25}
\end{figure*}

\begin{figure*}[ht]
    \centering
    \includesvg[width=0.8\textwidth]{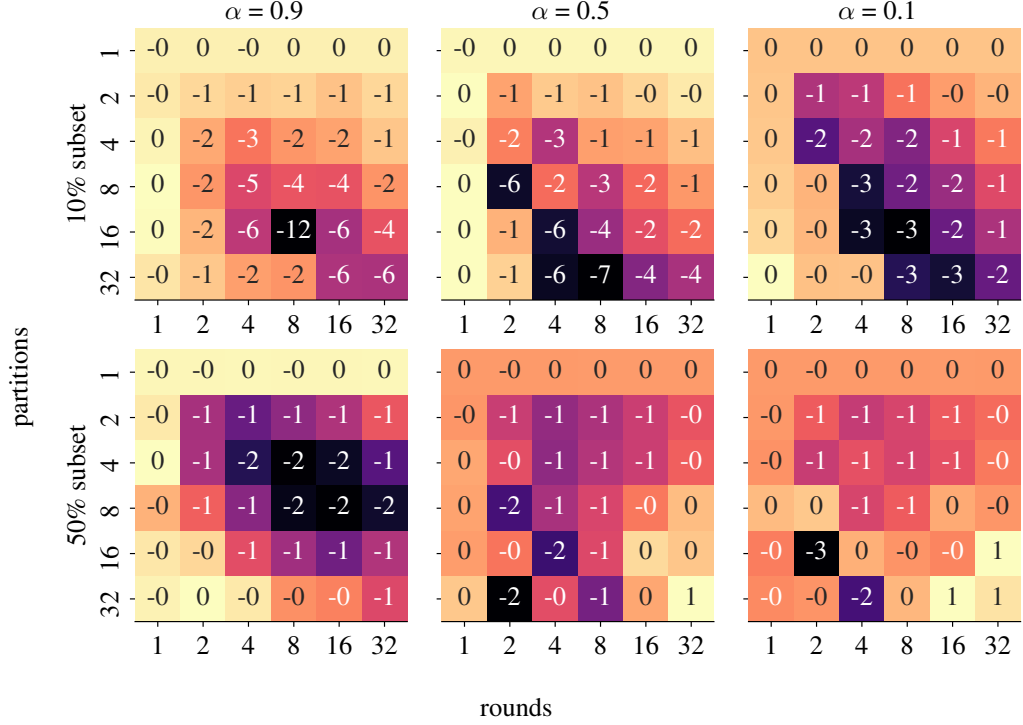}
    \caption{Difference in normalized score of \(\gamma = 1\) to \(\gamma = 0.75\) on ImageNet, depending on the subset size, the number of partitions, rounds, and \(\alpha\).} 
    \label{fig:imgnet-partition-rounds-noadap-gamma1}
\end{figure*}

\begin{figure*}[ht]
    \centering
    \includesvg[width=0.8\textwidth]{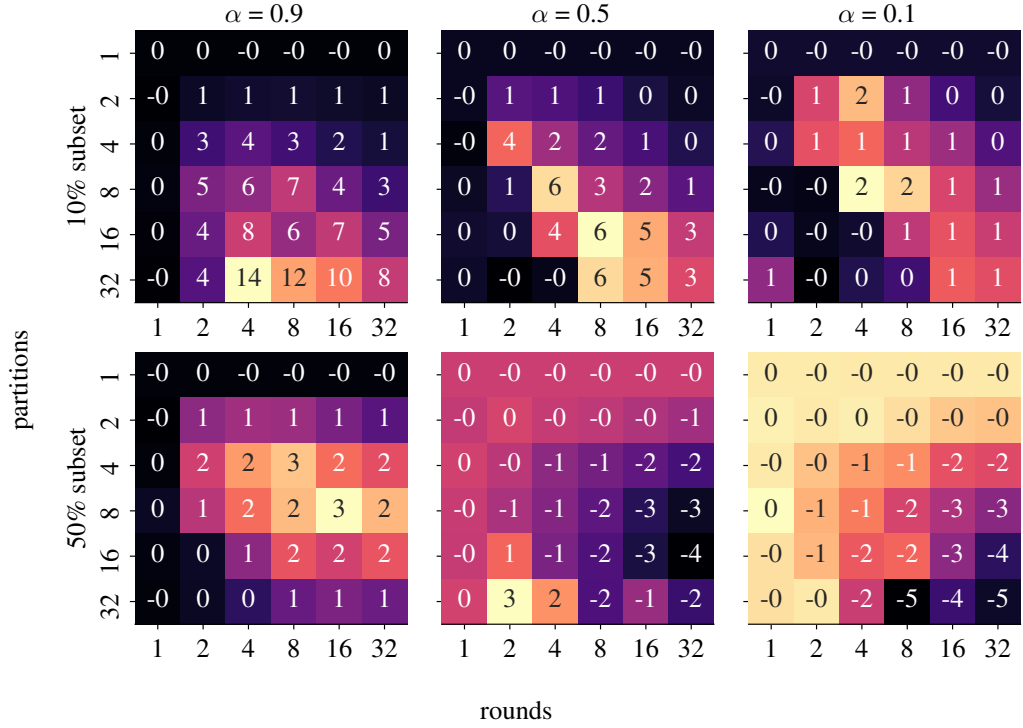}
    \caption{Difference in normalized score of \(\gamma = 0.5\) to \(\gamma = 0.75\) on ImageNet, depending on the subset size, the number of partitions, rounds, and \(\alpha\).} 
    \label{fig:imgnet-partition-rounds-noadap-gamma5}
\end{figure*}

\begin{figure*}[ht]
    \centering
    \includesvg[width=0.8\textwidth]{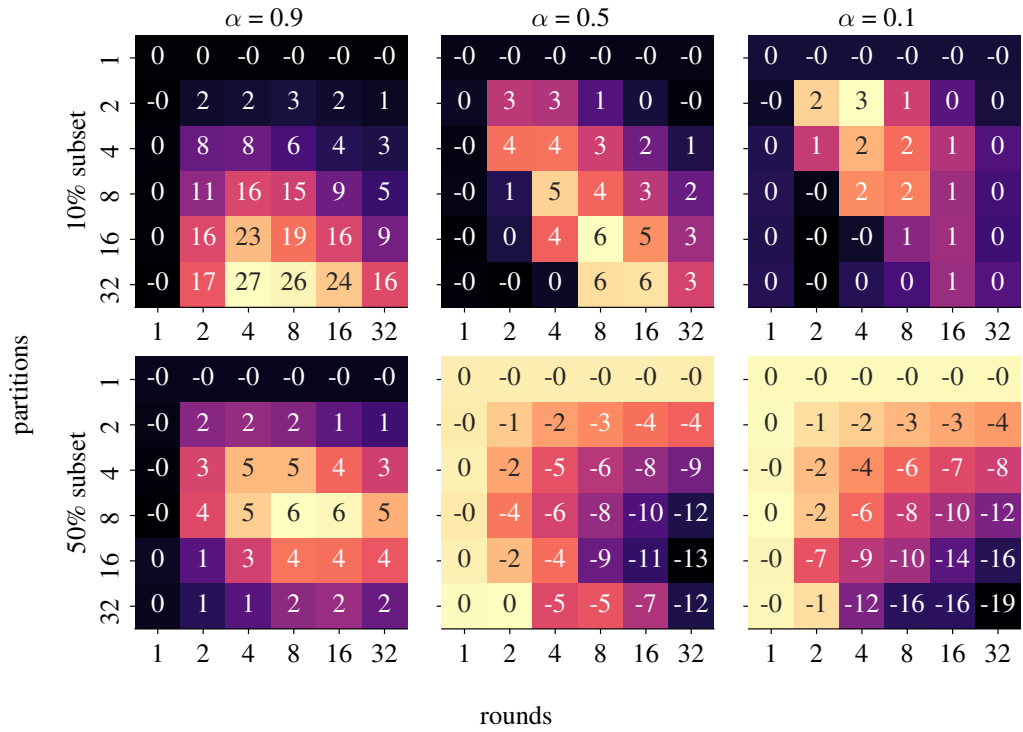}
    \caption{Difference in normalized score of \(\gamma = 0.25\) to \(\gamma = 0.75\) on ImageNet, depending on the subset size, the number of partitions, rounds, and \(\alpha\).} 
    \label{fig:imgnet-partition-rounds-noadap-gamma25}
\end{figure*}

We first evaluate \(\gamma\) without adaptive partitioning since adaptive partitioning is biased towards smaller values, as they allow for fewer partitions.
To ease the comparison, we investigate the difference of the normalized score to the base case of \(\gamma = 0.75\), i.e., positive values indicate a higher normalized score, and negative values indicate a lower negative score.
For CIFAR-100, the results can be found in~\Cref{fig:cifar-partition-rounds-noadap-gamma1,fig:cifar-partition-rounds-noadap-gamma5,fig:cifar-partition-rounds-noadap-gamma25} and for ImageNet, the results can be found in~\Cref{fig:imgnet-partition-rounds-noadap-gamma1,fig:imgnet-partition-rounds-noadap-gamma5,fig:imgnet-partition-rounds-noadap-gamma25}.
Note that decimal places are truncated in the plots.

For \(\gamma = 1\), i.e., when increasing the intermediate partition sizes, for both CIFAR and ImageNet, the scores are mostly similar or worse than \(\gamma = 0.75\), across all configurations.
There are just very few instances where scores are marginally higher.
At the same time, the compute and storage costs are higher due to the larger intermediate partition sizes.

For \(\gamma = 0.5\), i.e., when decreasing the intermediate partition sizes, we find an increase in scores in many scenarios.
This holds especially for \(\alpha = 0.9\), i.e., when utility is very important.
Intuitively, having smaller intermediate partitions forces the algorithm to make inclusion/exclusion decisions earlier.
For settings where utility is important, this is is beneficial since there is less noise by diverse data points in the selection process.
The benefit is higher the more partitions are used.
For \(\alpha = 0.1\) the inverse effect happens, as the more partitions are used, \(\gamma = 0.75\) performs better than \(\gamma = 0.5\).
Further lowering \(\gamma\) to \(0.25\), these effects are amplified.

\emph{
For larger values of \(\alpha\), smaller values of \(\gamma\) are preferred, while smaller values of \(\alpha\) benefit from larger intermediate partitions.
}

\FloatBarrier
\newpage
\section{Additional Figures (without bounding)}\label{app:figures}

In this section, we give the full versions and additional data omitted in the main body of the paper due to space constraints.
Due to the two-column layout, the figures appear only on the subsequent page.
\FloatBarrier

\begin{figure*}[ht]
    \centering
    \includesvg[width=0.9\textwidth]{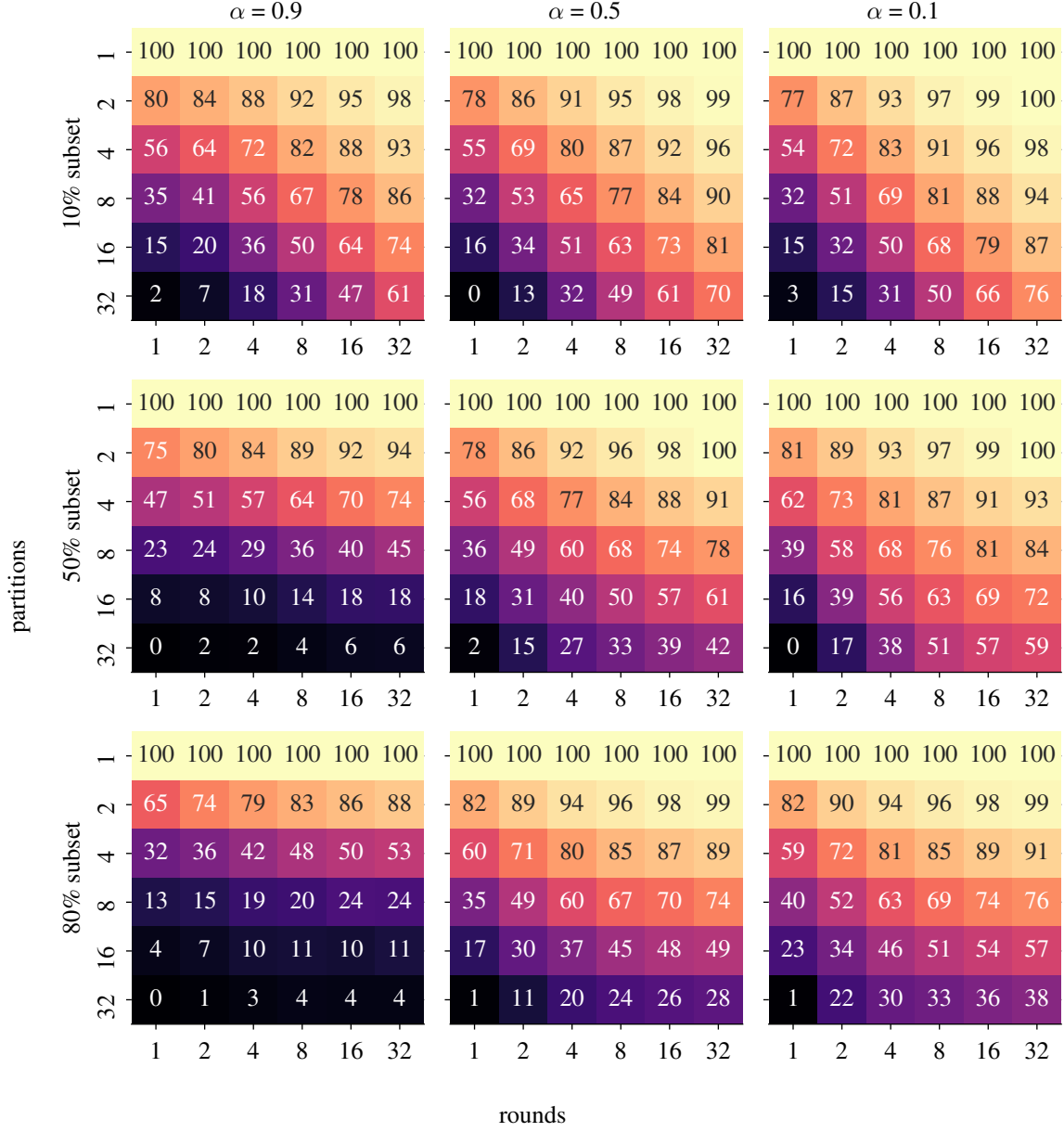}
    \caption{Full version of~\Cref{fig:parittions-rounds-cifar}. Normalized scores for finding subsets of CIFAR-100, depending on the subset size, the number of partitions, rounds, and \(\alpha\).} 
    \label{fig:partition-rounds-cifar-full}
\end{figure*}

\begin{figure*}[ht]
    \centering
    \includesvg[width=0.9\textwidth]{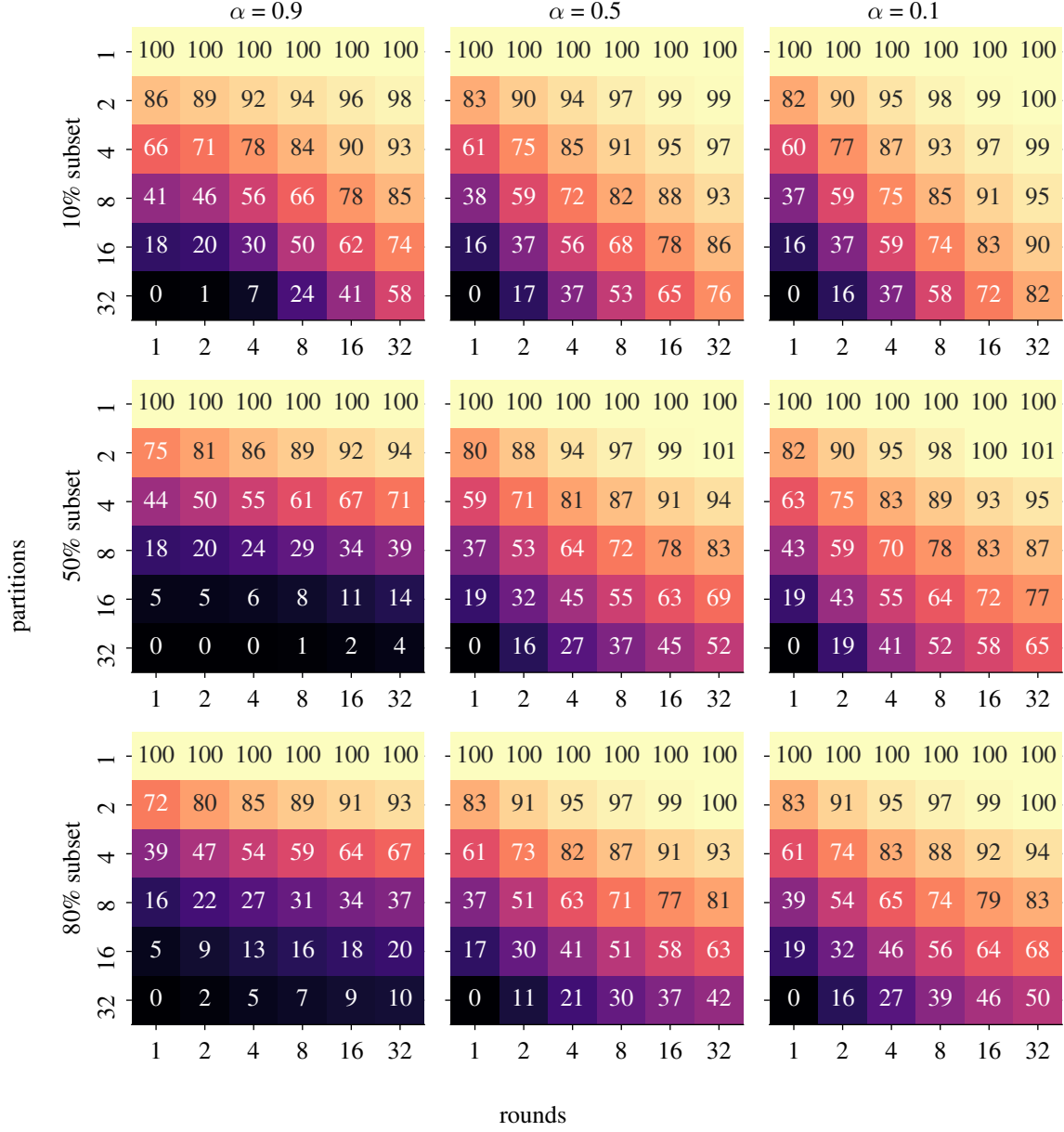}
    \caption{Normalized scores for finding subsets of ImageNet, depending on the subset size, the number of partitions, rounds, and \(\alpha\). } 
    \label{fig:partition-rounds-imagenet-full}
\end{figure*}

\begin{figure*}[ht]
    \centering
    \includesvg[width=0.9\textwidth]{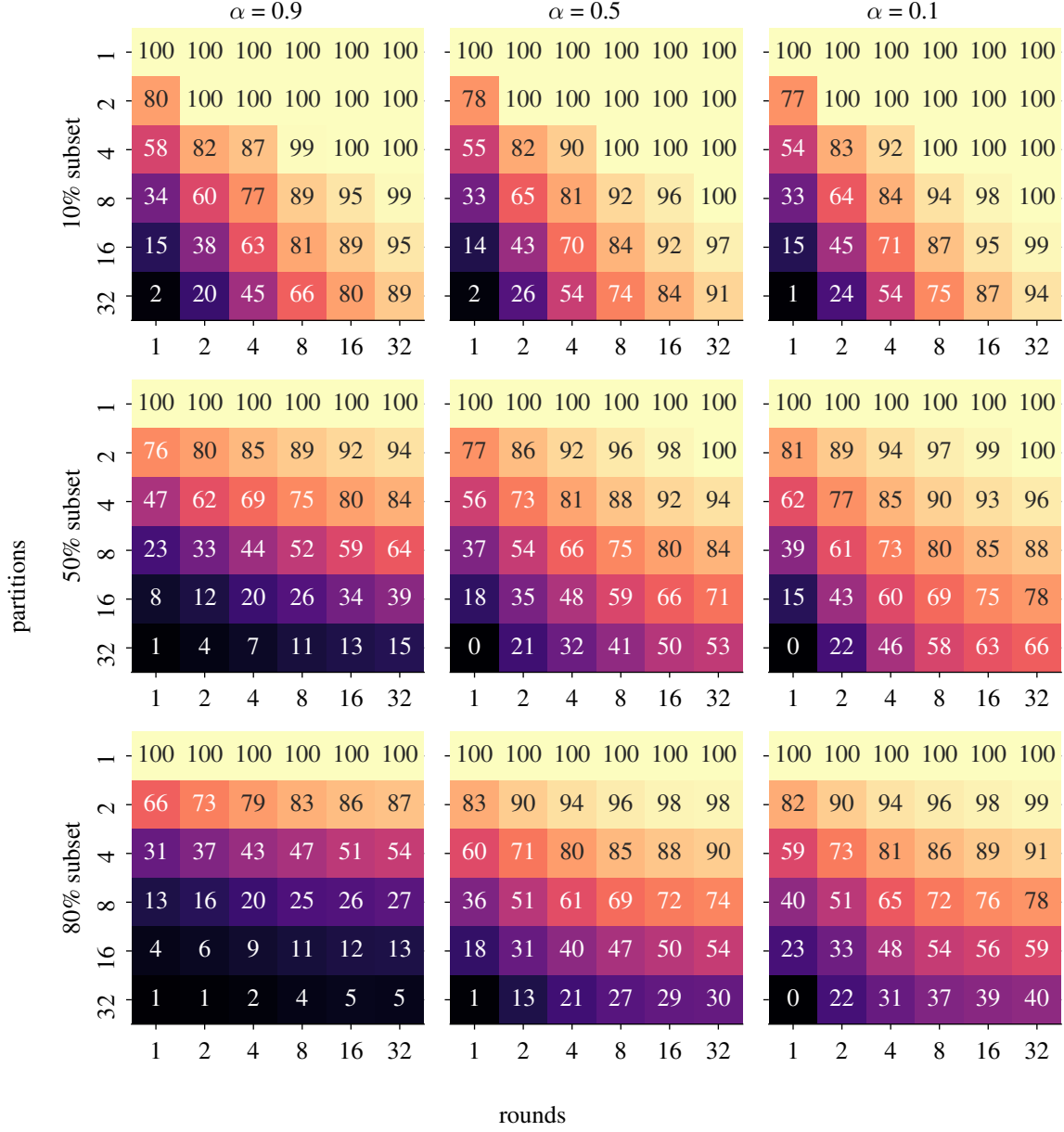}
    \caption{Full version of~\Cref{fig:adaptive-rounds-cifar}. Normalized scores for finding subsets of CIFAR-100, depending on the subset size, the number of partitions, rounds, and \(\alpha\), using adaptive partitioning.} 
    \label{fig:adaptive-rounds-cifar-full}
\end{figure*}

\begin{figure*}[ht]
    \centering
    \includesvg[width=0.9\textwidth]{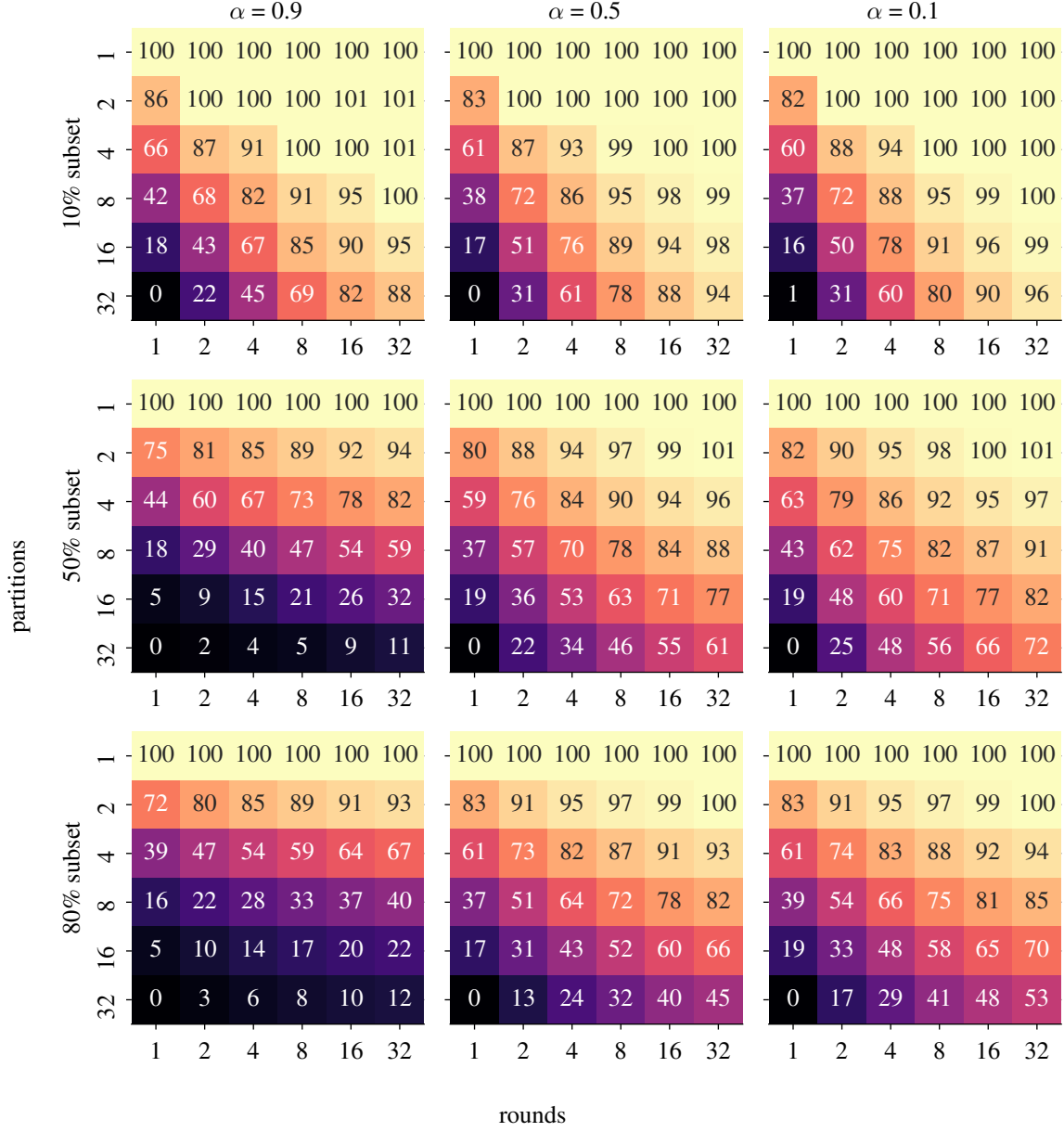}
    \caption{Normalized scores for finding subsets of ImageNet, depending on the subset size, the number of partitions, rounds, and \(\alpha\), using adaptive partitioning.} 
    \label{fig:adaptive-rounds-imagenet-full}
\end{figure*}

\FloatBarrier

\section{Additional Figures (with bounding)}\label{app:figures-bounding}
In this section, we give the full heatmaps of the five bounding configurations on ImageNet and CIFAR.
For statistics on the number of included points, refer to~\Cref{fig:bounding-tables}.
We only show the results for \(\alpha = 0.9\) since, as discussed in~\Cref{subsec:eval-bounding}, for other values of \(\alpha\), bounding does not include or exclude points.
Due to the two-column layout, the figures only appear on the subsequent page.

\begin{figure*}[ht]
    \begin{adjustbox}{addcode={\begin{minipage}{\width}}{\caption{%
    Normalized scores for finding subsets of CIFAR-100, depending on the subset size, the number of partitions, rounds, and type of bounding, using adaptive partitioning.}\end{minipage}},rotate=90,center}
    \centering
    \includesvg[width=1.2\textwidth]{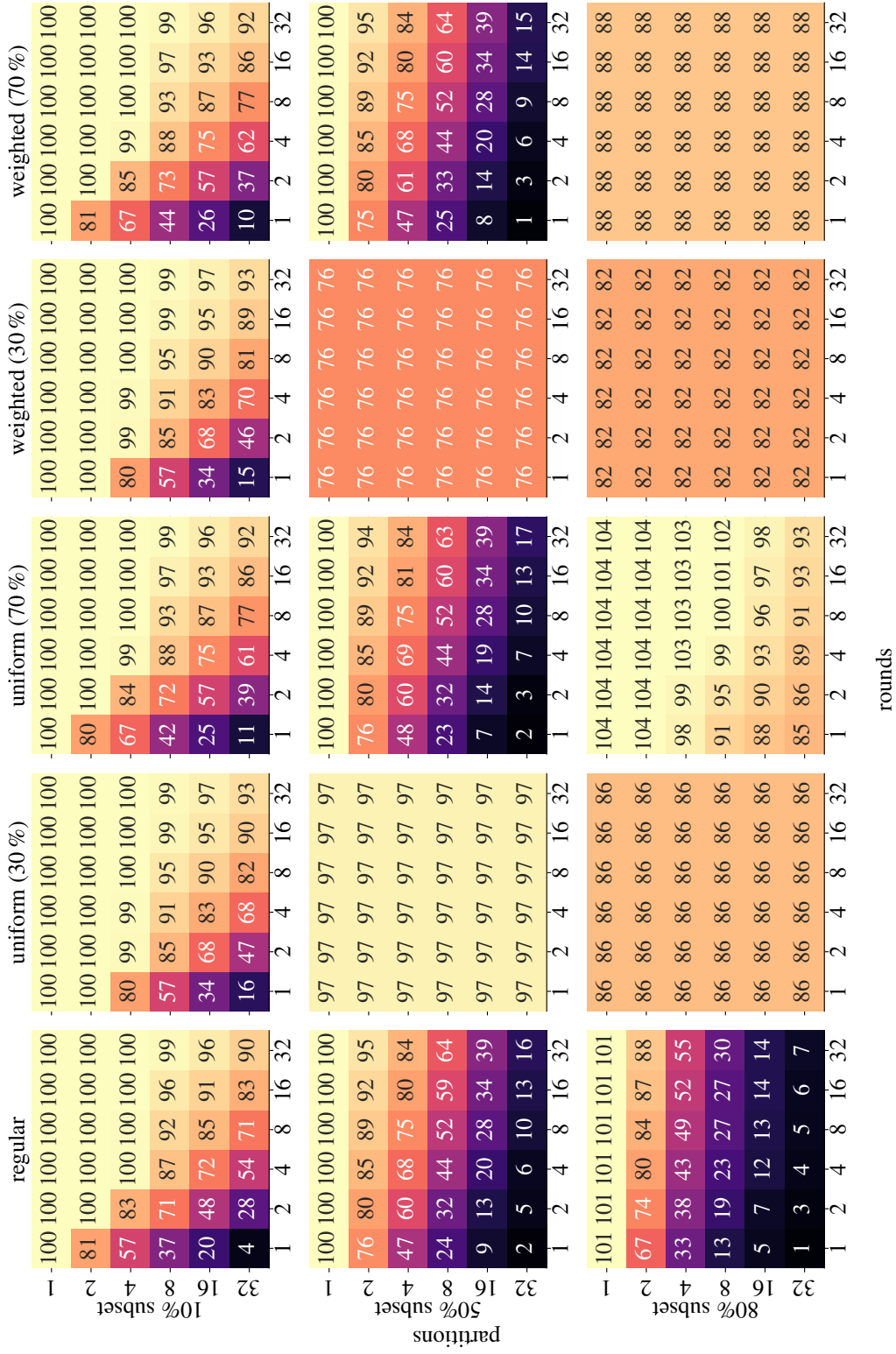}
    \end{adjustbox}
    \label{fig:bounding-cifar-full}
\end{figure*}

\begin{figure*}[ht]
    \begin{adjustbox}{addcode={\begin{minipage}{\width}}{\caption{%
    Normalized scores for finding subsets of ImageNet, depending on the subset size, the number of partitions, rounds, and type of bounding, using adaptive partitioning.}\end{minipage}},rotate=90,center}
    \centering
    \includesvg[width=1.2\textwidth]{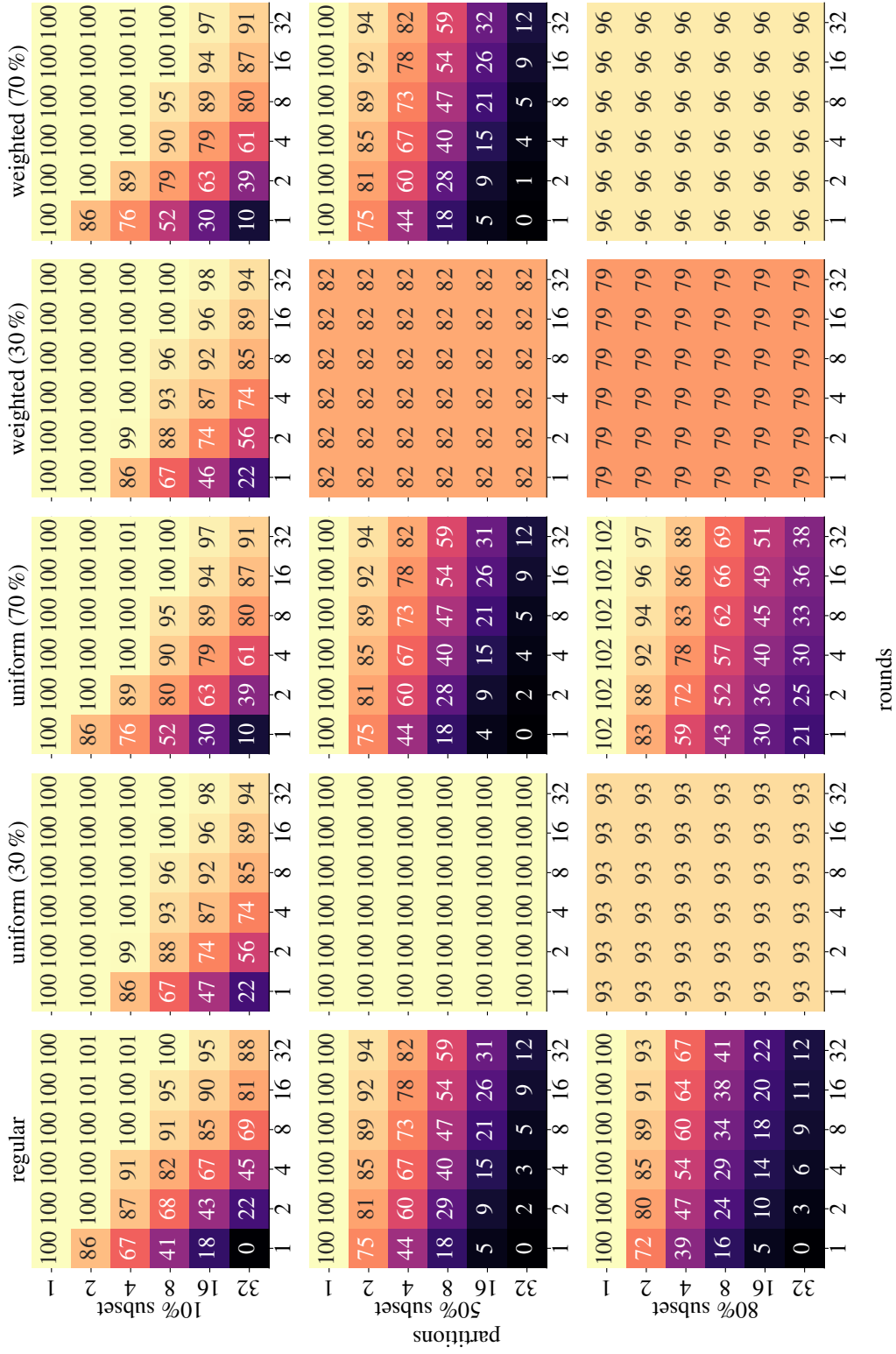}
    \end{adjustbox}
    \label{fig:bounding-imgnet-full}
\end{figure*}

\end{document}